\documentclass{article} 
\usepackage{iclr2027_conference,times}

\usepackage[utf8]{inputenc} 
\usepackage[T1]{fontenc}    
\usepackage{hyperref}       
\usepackage{url}            
\usepackage{booktabs}       
\usepackage{amsfonts}       
\usepackage{nicefrac}       
\usepackage{microtype}      
\usepackage{xcolor}         
\usepackage{graphicx}
\usepackage{amsmath}
\usepackage{multirow}
\usepackage{booktabs}
\usepackage{makecell}
\usepackage[breakable]{tcolorbox}
\tcbuselibrary{skins}
\usepackage{tabularx}
\usepackage{enumitem}
\usepackage{wrapfig}
\usepackage{hyperref}

\definecolor{myblue}{HTML}{4A90E2} 
\definecolor{myred}{HTML}{B22222} 

\title{The Illusion of Reasoning: Exposing Evasive Data Contamination in LLMs via Zero-CoT Truncation}

\author{
Yifan Lan,
Yuanpu Cao,
Hanyu Wang,
Lu Lin,
Jinghui Chen \\
The Pennsylvania State University \\
\texttt{\{yifanlan,ymc5533,hbw5365,lxl5598,jzc5917\}@psu.edu}
}

\newcommand{\ifcomments}{\iftrue}

\iclrfinalcopy

\begin{document}

\maketitle
\lhead{}

\begin{abstract}
Large language models (LLMs) have demonstrated impressive reasoning abilities across a wide range of tasks, but data contamination undermines the objective evaluation of these capabilities. This problem is further exacerbated by malicious model publishers who use evasive, or indirect, contamination strategies, such as paraphrasing benchmark data to evade existing detection methods and artificially boost leaderboard performance. Current approaches struggle to reliably detect such stealthy contamination.
In this work, we uncover a critical phenomenon: a model's generated reasoning steps actively mask its underlying memorization. Inspired by this, we propose the Zero-CoT Probe (ZCP), a novel black-box detection method that deliberately truncates the entire Chain-of-Thought (CoT) process to expose latent shortcut mappings. To further isolate memorization from the model's intrinsic problem-solving capabilities, ZCP compares the model's zero-CoT performance on the original benchmark against an isomorphically perturbed reference dataset. Furthermore, we introduce Contamination Confidence, a metric that quantifies both the likelihood and severity of contamination, moving beyond simple binary classifications. Extensive experiments on both previously identified contaminated models and specially fine-tuned contaminated models demonstrate that ZCP robustly detects both direct and evasive data contamination. The code for ZCP is accessible at \url{https://github.com/Yifan-Lan/zero-cot-probe}.
\end{abstract}

\section{Introduction}
\label{sec:intro}

Recent advances in Large Language Models (LLMs) \citep{gpt4,qwen3,llama3,gemini2-5} have yielded exceptional reasoning capabilities, further amplified by Chain-of-Thought (CoT) \citep{cot,auto-cot,o1}. As models achieve unprecedented performance across domains such as mathematics and code generation, rigorous evaluation via high-quality benchmarks \citep{gsm8k,gpqa,mmlu,swebench} becomes paramount. However, this evaluation paradigm is severely threatened by data contamination \citep{gpt3,gpt4,2024survey1,survey2025}, the intentional or inadvertent inclusion of benchmark data in training data. Contamination artificially inflates evaluation metrics, creating a dangerous illusion of capability. Consequently, it distorts developers' deployment decisions, and severely widens the gap between reported leaderboard scores and actual real-world utility for users.

While traditional detection methods exist, they face a formidable challenge in \textit{evasive} (indirect) data contamination \citep{evasive-data-contamination,llm-decontaminator,critic-verbatim-memorization}. Whether malicious publishers aggressively paraphrase benchmarks to game leaderboards, or models inadvertently ingest synthetic benchmark-like data, evasive scenarios severely alter exact phrasing. Consequently, current detectors relying on surface-level verbatim overlap fail entirely. Furthermore, the pervasive opacity of pre-training corpora renders direct inspection methods impossible.

\begin{figure}[tbhp!]
    \centering
    \includegraphics[width=0.78\textwidth]{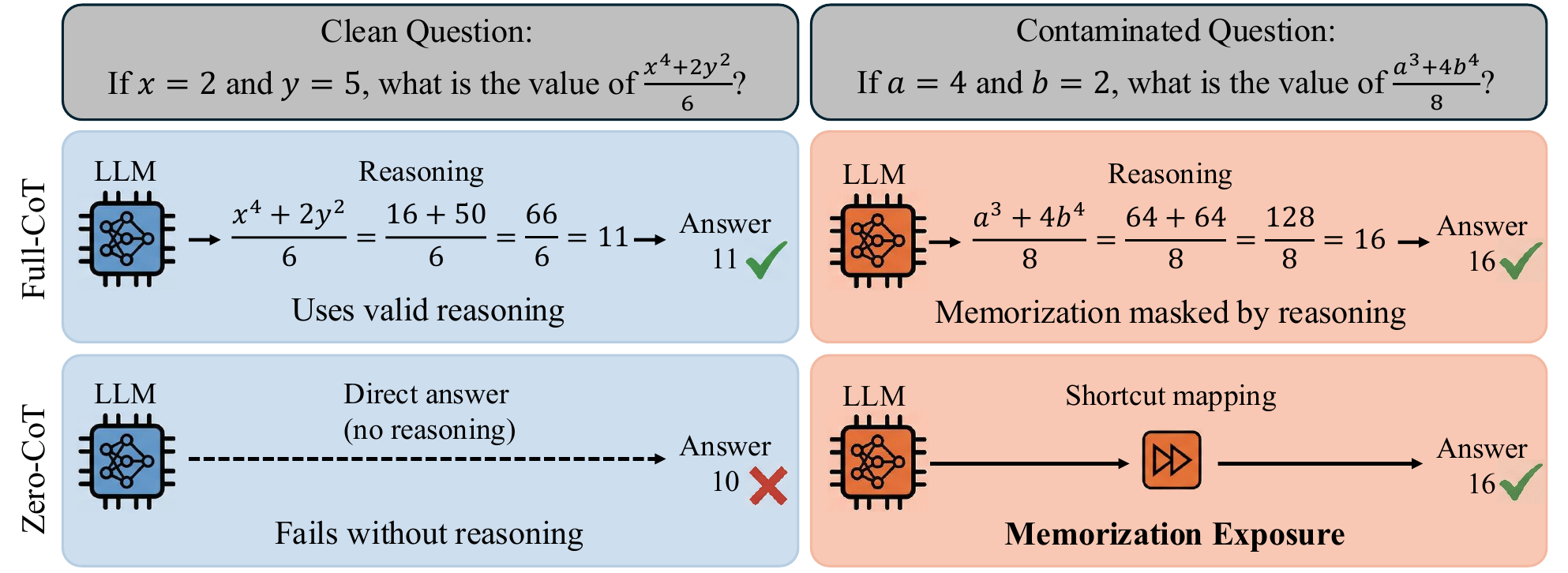}
    \caption{\textbf{Reasoning masks data contamination.} Under Full-CoT (Top), memorization is indistinguishable from genuine reasoning. Our Zero-CoT Probe (Bottom) forces the model to bypass intermediate reasoning. Consequently, the model fails on clean questions but still correctly answers contaminated ones via a learned shortcut mapping, thereby exposing the memorization.}
    \label{fig:cot_truncation}
\end{figure}

To address this, we introduce a novel method ZCP (Zero-CoT Probe) to detect evasive contamination by leveraging the Chain-of-Thought (CoT) capabilities of LLMs. We observe that if a model has been trained on a specific dataset, even a paraphrased one, it establishes a direct, shortcut mapping from the semantics of the question $x_i$ to the answer $y_i$, making it significantly more likely to generate the correct final answer without CoT, as illustrated in Figure~\ref{fig:cot_truncation}. Specifically, our method isolates this memorization by truncating the CoT and forcing the model to generate the final answer directly. To further exclude the possibility that the model possesses some ``superpower'' (an informal shorthand we use throughout for the model's intrinsic, non-memorization-based ability to answer complex questions without explicit reasoning), we compare its zero-CoT performance on the original benchmark against a cleaned reference dataset. A severe performance drop on the reference data explicitly exposes contamination. ZCP is a practical solution that does not require access to the LLM's training data or parameters.

The main contributions of this paper are as follows:
\begin{itemize}[leftmargin=2em]
\setlength\itemsep{0em}
    \item We uncover that reasoning can actively mask underlying memorization. Inspired by this, we propose a novel black-box method that truncates CoT and utilizes isomorphically perturbed reference data to robustly detect both direct and evasive contamination.
    \item We introduce Contamination Confidence, a new statistical metric to quantify the benchmark-level data contamination severity, advancing beyond simple binary detection results.
    \item We systematically evaluate the real-world data contamination levels of prominent closed-source and open-source models, revealing the broad existence of data contamination.
\end{itemize}

\vspace{-10pt}
\section{Related Work}
\textbf{Data Contamination.} Data contamination occurs when evaluation benchmarks are included in a model's training corpus, artificially inflating performance metrics on these benchmarks \citep{gpt3,gpt4,2024survey1}. While existing methods can detect standard verbatim contamination \citep{quantify-data-leakage,data-reconstruction,investigating-data-contamination,perplexity,shuffle-multiple-choice,member-inference}, they struggle against evasive (or indirect) contamination. This stealthy variant occurs when benchmarks are aggressively paraphrased to manipulate leaderboards \citep{evasive-data-contamination,llm-decontaminator,critic-verbatim-memorization}, or inadvertently ingested via synthetic samples during knowledge distillation \citep{indirect-data-contamination-by-llm}.

Existing defenses against evasive contamination remain severely limited. Probabilistic detection \citep{mink-prob} falls short under heavy paraphrasing \citep{evasive-data-contamination}. \citet{llm-decontaminator} proposed a robust two-stage similarity approach, yet it impractically requires full access to the suspect model's pre-training data. Alternatively, \citet{ccd-method} detect anomalies via low output variance, assuming memorization strictly induces determinism. However, this low-variance assumption fails for modern LLMs trained via Reinforcement Learning (e.g., GRPO \citep{GRPO}), which explicitly incentivizes diverse reasoning trajectories. Furthermore, their evaluation is heavily biased by rigid coding tasks, where strict syntax naturally restricts variance, undermining the method's generalizability to broader reasoning domains.

\textbf{Research on CoT.} Beyond enhancing task performance \citep{cot,auto-cot,o1}, Chain-of-Thought (CoT) interventions are increasingly used to probe LLM internals. For instance, prior works have manipulated CoT to assess reasoning faithfulness \citep{faithfulness-of-cot,faithfulness-of-cot2} or truncated it to analyze reward hacking \citep{reward-hacking}. Building on this analytical paradigm, we force LLMs to bypass reasoning entirely (zero-CoT) to investigate data contamination. Our core intuition is that memorization establishes a latent shortcut mapping, allowing models to produce correct answers without rigorous reasoning. By truncating CoT, we neutralize reasoning as a confounder, thereby directly exposing these memorized shortcuts when compared against performance on  reference data.

\vspace{-10pt}
\section{Method}
\vspace{-5pt}

In this section, we first formally define the problem of evasive data contamination in Section~\ref{sec:problem-formulation}. We then analyze the inherent limitations of existing detection methods in Section~\ref{sec:limitations-of-existing-detection}, which naturally motivates our proposed detection framework detailed in Sections~\ref{sec:cot_truncation} through \ref{section:contamination-confidence}.

\vspace{-5pt}
\subsection{Problem Formulation}
\label{sec:problem-formulation}

Let $M$ be a target Large Language Model and $D_{eval} = \{(x_i, y_i)\}_{i=1}^N$ be an evaluation benchmark, where $x_i$ denotes a question requiring multi-step reasoning and $y_i$ is the ground-truth answer. 

\textbf{Standard Data Contamination} occurs when the benchmark data is explicitly included in the model's pre-training or fine-tuning corpus $D_{train}$. In this case, $(x_i, y_i) \in D_{train}$, allowing the model to directly memorize the exact string sequences.

\textbf{Evasive (Indirect) Data Contamination} occurs when the evaluation data is paraphrased or syntactically altered before being included in the training corpus. This arises intentionally when a malicious publisher obfuscates the benchmark data to bypass detection and inflate leaderboard rankings. It can also happen inadvertently during knowledge distillation when a model is trained on synthetic samples generated by other LLMs that closely mirror benchmark data, or when web-scraped training corpora include online discussions that rephrase benchmark questions. In either scenario, the model is trained on a modified dataset $D'_{eval} = \{(x'_i, y'_i)\}_{i=1}^N$, where $x'_i \neq x_i$ at the surface level, but the semantic meaning, underlying logical structure, and ground-truth answer ($y'_i = y_i$) remain identical.



The goal of our work is to design a detection function $f(M, D_{eval}) = \mathcal{C}$, where \textit{Contamination Confidence} score $\mathcal{C} \in [0.5, 1]$ quantifies the extent to which $M$ has memorized $D_{eval}$ (either directly or evasively). In this formulation, a baseline score of $\mathcal{C} = 0.5$ denotes no statistical evidence of contamination (i.e., the result is indistinguishable from random variance), whereas $\mathcal{C} \to 1.0$ indicates definitive memorization. 
Crucially, this function operates in a strictly black-box setting: it does not require access to the training corpus $D_{train}$ or the target model's internal parameters, which are aligned with practical scenarios.

\vspace{-5pt}
\subsection{Limitations of Existing Detection Methods in Evasive Scenarios}
\label{sec:limitations-of-existing-detection}


Existing contamination detection methods fail against evasive data contamination for two main reasons. First, methods measuring n-gram overlap or embedding similarity \citep{gpt3,llm-decontaminator} impractically require access to the target model's training corpus ($D_{train}$), a transparency rarely offered by malicious publishers.


For black-box auditing (without access to training data), current paradigms strictly rely on verbatim, token-level memorization, making them easily exploitable. Likelihood-based metrics (e.g., perplexity or DPCC) \citep{mink-prob,DPCC,perplexity} assume the exact original tokens of $(x_i, y_i)$ yield abnormally high probabilities. However, evasive data contamination alters these exact lexical sequences, rendering the metrics ineffective.
DPCC is one of these methods, calculating the RMIA metric (the proportion that the loss of the original sample is larger than augmented ones) for each sample. If the proportion of samples with an RMIA score below 0.1 exceeds a threshold of 0.85, the benchmark is classified as contaminated. We present the performance of DPCC in Table~\ref{tab:DPCC_results}. Although all the scores are below the threshold of 0.85, some in the original scenarios like GSM8K and MATH on Qwen2.5-Math are comparatively high. So, if adjusting the threshold, the detection on these scenarios may succeed. However, scores of paraphrased datasets are always much lower than original datasets, implying the failure of DPCC on evasive data contamination.

\begin{table*}[bthp]
    \centering
    \small 
    \renewcommand{\arraystretch}{1.1} 
    
    \begin{minipage}[b]{0.49\textwidth}
        \centering
        \caption{Scores of DPCC on original and paraphrased datasets on Qwen 2.5-Math and DeepSeek-Math.}
        \label{tab:DPCC_results}
        
        \begin{tabularx}{\linewidth}{@{} l *{4}{>{\centering\arraybackslash}X} @{}}
            \toprule
            \multirow{2}{*}{\textbf{Version}} & \multicolumn{2}{c}{\textbf{Qwen2.5-Math}} & \multicolumn{2}{c}{\textbf{DeepSeek-Math}} \\
            \cmidrule(lr){2-3} \cmidrule(lr){4-5}
            & GSM8K & MATH & GSM8K & MATH \\
            \midrule
            Original    & 0.420 & 0.730 & 0.052 & 0.366 \\
            Paraphrased & 0.062 & 0.191 & 0.028 & 0.104 \\
            \bottomrule
        \end{tabularx}
    \end{minipage}\hfill 
    \begin{minipage}[b]{0.49\textwidth}
        \centering
        \caption{Performance of the data reconstruction detection method on original and paraphrased contaminated datasets on Qwen2.5-Math.}
        \label{tab:reconstruction_failure}
        
        \begin{tabularx}{\linewidth}{@{} l *{4}{>{\centering\arraybackslash}X} @{}}
            \toprule
            \multirow{2}{*}{\textbf{Version}} & \multicolumn{2}{c}{\textbf{ROUGE-L}} & \multicolumn{2}{c}{\textbf{Accuracy}} \\
            \cmidrule(lr){2-3} \cmidrule(lr){4-5}
            & GSM8K & MATH & GSM8K & MATH \\
            \midrule
            Original    & 0.551 & 0.621 & 0.398 & 0.386 \\
            Paraphrased & 0.213 & 0.267 & 0.176 & 0.191 \\
            \bottomrule
        \end{tabularx}
    \end{minipage}
    \vspace{-10pt}
\end{table*}

Another paradigm detects contamination via data reconstruction (sequence completion) \citep{qwen-math-data-contamination,data-completion1,data-completion2}. We evaluated this by providing a 40\% question prefix and sampling 16 completions, measuring the maximum ROUGE-L overlap and pass@16 accuracy. As demonstrated in Table~\ref{tab:reconstruction_failure}, while effective on original data (standard data contamination), reconstruction performance plummets on paraphrased data (evasive contamination). This failure stems from its strict reliance on verbatim, token-level memorization, which is easily destroyed by the syntactic and vocabulary alterations in paraphrased datasets.

Similarly, ``guided instruction'' \citep{data-reconstruction} attempts reconstruction by appending inadvertently leaked dataset metadata (e.g., partition names) to the prefix. They assume that the associated dataset name and the partition are inadvertently leaked during the pre-training stage. However, malicious evasive contamination typically occurs during fine-tuning \citep{evasive-data-contamination,ccd-method}, and publishers can easily strip or obfuscate such metadata, rendering the method ineffective.

These vulnerabilities highlight a critical blind spot: they rely on easily obfuscated surface-level features. To expose true evasive data contamination, we must probe deeper into the model's learned mappings and bypass the confounding intermediate reasoning chain entirely. By enforcing a zero-CoT generation setting, we neutralize complex reasoning noise, forcing the underlying memorization to reveal itself through the direct mappings from the question $x_i$ to the final answer $y_i$.

\vspace{-6pt}
\subsection{Neutralizing Reasoning via CoT Truncation}
\label{sec:cot_truncation}

\begin{wrapfigure}{r}{0.55\textwidth}
    \centering
    \vspace{-15pt}
    \includegraphics[width=0.99\linewidth]{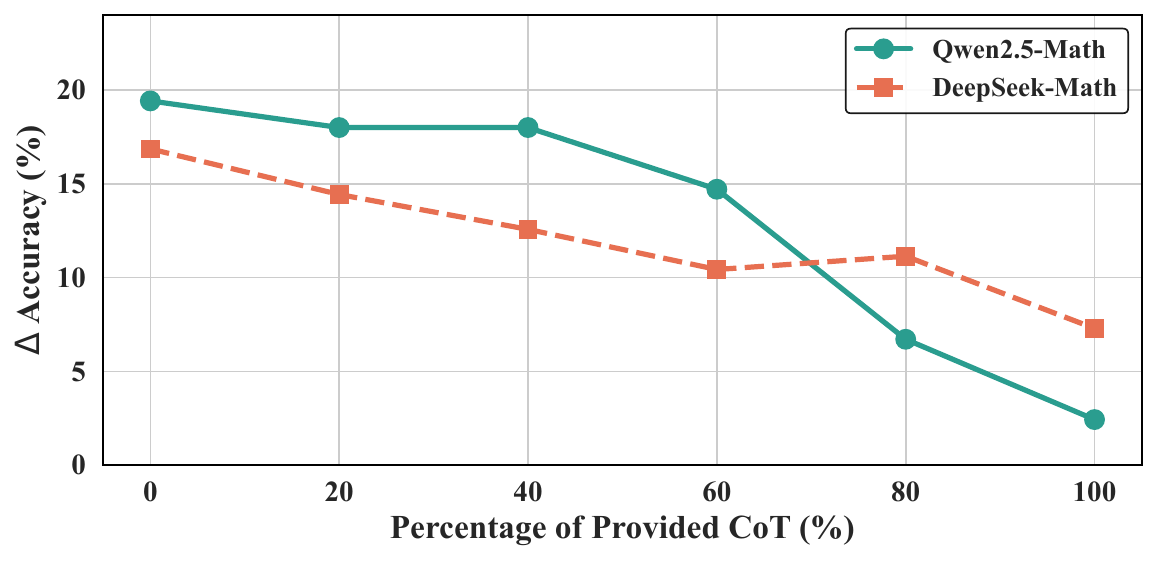}
    \vspace{-15pt}
    \caption{The accuracy gap ($\Delta$) between contaminated and clean questions across varying CoT percentages. As the reasoning chain is systematically omitted, the gap widens drastically. 
    }
    \label{fig:masking_curve}
    \vspace{-15pt}
\end{wrapfigure}
In standard generation processes, LLMs solve complex problems via a Full-CoT (default) generation setting. Given an input question $x_i$, the model first generates an intermediate reasoning chain $\hat{c}_i$, and then produces the final answer $\hat{y}_i$. The probability of generating the correct answer is thus heavily conditioned on the reasoning steps.
For challenging tasks, models rely heavily on generating a valid and rigorous reasoning path $\hat{c}_i$ to get a high accuracy. 


However, we hypothesize that if a model has memorized the dataset during training, it develops a latent shortcut mapping directly from the semantics of $x_i$ to $y_i$. Consequently, when the intermediate reasoning chain $\hat{c}_i$ is omitted, the model exhibits a significantly higher probability of producing the correct final answer for a contaminated question compared to an unseen clean question. We provide direct empirical evidence for this latent shortcut in Figure~\ref{fig:masking_curve}: as the provided reasoning chain is systematically truncated (approaching 0\%), the accuracy gap between contaminated and clean questions widens drastically, confirming the model's reliance on these direct mappings when reasoning is disabled.

Motivated by these findings, we can deliberately truncate the CoT entirely to neutralize the influence of the reasoning factor, thereby unmasking the underlying memorization. This intervention forces the model to rely on the remaining two factors to produce a correct output: either the memorization of the dataset, or an intrinsic ``superpower'' to solve complex problems without intermediate steps. Crucially, without this truncation, the model's reasoning ability actively masks its memorization, acting as a severe confounder in contamination detection, as conceptually illustrated in Figure~\ref{fig:cot_truncation}. We further validate this masking effect and the absolute necessity of CoT truncation in Section~\ref{sec:influence-of-reasoning}.


To exploit this, we enforce a Zero-CoT generation setting. Given a question $x_i$, we construct a forced prompt $\hat{x}_i$ that enforces the model to output the final answer immediately without CoT. The precise construction of $\hat{x}_i$ depends on model accessibility. For open-weight models (e.g., Qwen), we append the prefix \texttt{"The final answer is: \textbackslash[ \textbackslash boxed\{"} to the beginning of model response, forcing it to complete the final answer seamlessly. For closed-source models (e.g., the GPT series) where response prefixes cannot be explicitly pre-filled, we construct $\hat{x}_i$ by adding a strict instruction to the end of the user query: \texttt{"Please ONLY put your final answer within \textbackslash boxed\{\} directly without any other content before or after it (e.g., reasoning or explanation)"}. We observe that these forced prompts $\hat{x}_i$ consistently succeed in forcing models to output final answers directly.


\subsection{Performance Metric}
\label{sec:performance-metric}

We then evaluate the model $M$'s performance under this Zero-CoT constraint. Let the ground-truth $y_i$ consist of a sequence of $K$ tokens $(t_1, t_2, \dots, t_K)$. We define $S(M, x_i)$ as the performance metric on $x_i$. Because we want to do benchmark-level detection, we calculate the average performance metric on the whole dataset $D_{eval}$, denoted as $S(M, D_{eval})$. 
We employ four distinct metrics in our experiments to capture both discrete correctness and continuous probability distributions:

\vspace{-5pt}
\begin{itemize}[leftmargin=2em]
\setlength\itemsep{0em}
    \item \textbf{Accuracy ($Acc$):} A discrete metric ($S_{acc}(M, x_i) \in \{0, 1\}$) indicating whether the model's generated final answer $\hat{y}_i$ under the zero-CoT setting matches the ground truth $y_i$.
    \item \textbf{Consistency ($Con$):} A discrete metric ($S_{con}(M, x_i) \in \{0, 1\}$) indicating whether the zero-CoT final answer aligns with the answer generated under the default full-CoT setting. This measures the model's reliance on its reasoning chain.
    \item \textbf{First Token Probability ($\mathcal{P}_{first}$):} The generation probability of the very first token of the ground-truth answer, conditioned on the truncated prompt. This captures the model's immediate reflex to output the memorized answer:
    $\mathcal{P}_{first} = P(t_1 \mid \hat{x}_i)$
    \item \textbf{All Token Probability ($\mathcal{P}_{all}$):} The geometric mean of the token probabilities over the entire ground-truth answer, computed via teacher forcing. This metric normalizes for answer length and reflects the overall probability of generating the exact memorized sequence:
    $\mathcal{P}_{all} = \exp\left(\frac{1}{K} \sum_{k=1}^K \log P(t_k \mid \hat{x}_i, t_{<k})\right)$
\end{itemize}
\vspace{-5pt}

Rather than aggregating these metrics, we retain them individually to establish a versatile, multi-tiered auditing framework:
(1) \textbf{Logit-based metrics} ($\mathcal{P}_{first}$, $\mathcal{P}_{all}$): Require access to internal probability distributions, providing granular signals ideal for open-weight models.
(2) \textbf{Output-only metrics} ($Acc$, $Con$): Rely solely on the final generated text, scaling seamlessly to API-gated systems. Notably, $Con$ uniquely operates without ground-truth labels, further relaxing data access constraints. 
Metric robustness is further analyzed in Appendix~\ref{sec:ablation-datasize}.

\vspace{-5pt}
\subsection{Isolating Memorization via Reference Data}
\label{sec:reference}

While neutralizing the reasoning factor via CoT truncation is a crucial first step, it does not fully isolate memorization. High zero-CoT performance could stem from either true memorization or the model's intrinsic capability to perform complex internal calculations without emitting observable reasoning steps. To decouple these two factors and exclude the influence of this intrinsic ``superpower'', we introduce a control group by constructing a cleaned dataset as reference, denoted as $\tilde{D}_{eval}$. The zero-CoT performance on this reference data serves as a baseline of the ``superpower''. While establishing a reference group is a standard paradigm in data contamination detection, prior works typically rely on a clean reference model \citep{perplexity,reference-model2,reference-model3}. However, obtaining a guaranteed clean reference LLM is highly impractical, given the prohibitive computational costs of training from scratch and the opacity of existing pre-training corpora. 


To ensure $\tilde{D}_{eval}$ accurately isolates the baseline of the model's ``superpower'', it must perfectly mirror the difficulty and reasoning depth of the original benchmark $D_{eval}$. We observe that quantitative elements are prevalent in most complex reasoning tasks. Leveraging this, we apply an isomorphic perturbation strategy: we systematically alter the numerical values within the original question $x_i$ (maintaining the same order of magnitude) and paraphrase the textual context, while strictly retaining the original logical structure and reasoning depth, as illustrated by the case study in Table~\ref{tab:data_case_study}. Numerical perturbation has also been used to test reasoning robustness \citep{gsm-symbolic,gsm-plus,math-perturb}; we pair it with zero-CoT truncation to isolate memorization instead. This yields a semantically novel yet structurally isomorphic question $\tilde{x}_i$, with updated reasoning path $\tilde{c}_i$ and ground-truth answer $\tilde{y}_i$. Consequently, the cognitive load required to solve $x_i$ and $\tilde{x}_i$ remains entirely equivalent \footnote{Model's performance on the original and reference datasets remains statistically identical when evaluating under a standard Full-CoT setting, verifying the equivalent difficulty. Details are provided in Appendix~\ref{sec:influence-of-reasoning}.}. To execute this at scale, we design an automated, multi-model generation pipeline to synthesize and validate the reference dataset $\tilde{D}_{eval}$, as illustrated in Appendix \ref{appendix:multi-model-system}. We further verify the quality of $\tilde{D}_{eval}$ with human judgment in Appendix~\ref{appendix:human-eval}.



By comparing the zero-CoT performance on the original dataset $S(M, D_{eval})$ against the cleaned reference dataset $S(M, \tilde{D}_{eval})$, we systematically decouple memorization from ``superpower''. Equivalent performance ($S(M, D_{eval}) \approx S(M, \tilde{D}_{eval})$) implies that the model genuinely possesses intrinsic ``superpowers,'' indicating a clean dataset. Conversely, a statistically significant gap ($S(M, D_{eval}) > S(M, \tilde{D}_{eval})$) reveals that the model successfully answers the original questions but fails on logically identical reference questions of the same difficulty. This asymmetric degradation exposes data contamination, as the model's memorized shortcut mappings are effectively broken by the novel variable values introduced in $\tilde{D}_{eval}$. Since $\tilde{D}_{eval}$ is not released publicly, this comparison remains robust even against an adaptive adversary who anticipates our evaluation protocol; we analyze this robustness in Appendix~\ref{appendix:adaptive_contamination}. Our construction targets numerical reasoning tasks, but ZCP itself only requires CoT truncation and comparison against a valid reference dataset, so extending the construction strategy would extend ZCP's applicability equally.




\begin{table*}[tbhp]
    \centering
    \vspace{-10pt}
    \caption{A case study comparing the Original, Paraphrased, and Reference Cleaned data. The \textcolor{myblue}{blue text} indicates semantic paraphrasing that strictly preserves the mathematical logic, while the \textbf{\textcolor{myred}{red text}} highlights the isomorphic numerical perturbations that alter the final answer.}
    \label{tab:data_case_study}
    \renewcommand{\arraystretch}{1.1} 
    \setlength{\tabcolsep}{4pt}
    \resizebox{0.90\textwidth}{!}{
    \begin{tabular}{>{\raggedright\arraybackslash}p{0.14\textwidth} p{0.85\textwidth} c}
        \toprule
        \textbf{Data Type} & \textbf{Question} & \textbf{Answer} \\
        \midrule
        
        \textbf{Original} \newline $(D_{eval})$ & 
        Jack has a stack of books that is 12 inches thick. He knows from experience that 80 pages is one inch thick. If he has 6 books, how many pages is each one on average? & 
        160 \\
        
        \midrule
        
        \textbf{Paraphrased} \newline $(D'_{eval})$ & 
        \textcolor{myblue}{Maria} has a \textcolor{myblue}{pile of tomes whose combined spine thickness measures} 12 inches. \textcolor{myblue}{She} knows \textcolor{myblue}{that each inch corresponds to} 80 pages. If \textcolor{myblue}{her collection consists of} 6 \textcolor{myblue}{separate volumes}, \textcolor{myblue}{what is the mean number of} pages \textcolor{myblue}{in each volume}? & 
        160 \\
        
        \midrule
        
        \textbf{Cleaned} \newline ($\tilde{D}_{eval}$) & 
        \textcolor{myblue}{Emily} has a \textcolor{myblue}{collection of notebooks stacked to a height of} \textbf{\textcolor{myred}{15}} inches. \textcolor{myblue}{She has learned that} \textbf{\textcolor{myred}{90}} pages \textcolor{myblue}{make up} one inch \textcolor{myblue}{of thickness}. If \textcolor{myblue}{she owns} \textbf{\textcolor{myred}{5}} \textcolor{myblue}{notebooks}, how many pages \textcolor{myblue}{does each one contain} on average? & 
        \textbf{\textcolor{myred}{270}} \\
        
        \bottomrule
    \end{tabular}
    }
    \vspace{-10pt}
\end{table*}

\subsection{Quantifying Contamination Confidence}
\label{section:contamination-confidence}


Having isolated the memorization factor, we now formalize the calculation of the final \textit{Contamination Confidence} score, denoted as $\mathcal{C}_{cont}$. Prior works typically adopt a binary ``clean vs. contaminated'' classification, which fundamentally fails to capture the continuous spectrum of contamination caused by varying training exposure frequencies \citep{ccd-method,evasive-data-contamination} and leakage proportions \citep{survey3}. To move beyond this binary view, we adopt a statistical framework that calibrates frequentist $p$-values into a continuous, bound-based contamination evidence score.

First, we quantify the significance of the performance gap between $D_{eval}$ and $\tilde{D}_{eval}$ via a one-sided test, where the null hypothesis ($H_0$) posits no contamination ($S(M, D_{eval}) \le S(M, \tilde{D}_{eval})$) and the alternative hypothesis ($H_1$) posits that the dataset is contaminated. To robustly handle limited benchmark sample sizes without invoking rigid parametric assumptions \citep{t-test}, we employ a non-parametric bootstrap test \citep{bootstrap-test} with 10,000 resampling iterations for continuous metrics ($\mathcal{P}_{first}$, $\mathcal{P}_{all}$), and McNemar's test \citep{mcnemar-test} for discrete metrics ($Acc$, $Con$). 

Rather than thresholding this $p$-value for binary classification like existing benchmark-level contamination detection methods \citep{data-reconstruction,shuffle-multiple-choice}, we calibrate it into a bound-based contamination evidence score using the calibration of \citep{bayes-factor,sellke-calibration}. Assuming a proper $p$-value (uniformly distributed under $H_0$), the upper bound of the Bayes Factor ($\text{BF}_{10}$)—which quantifies the maximum evidence favoring contamination ($H_1$) over $H_0$—is formulated as:
\begin{equation}
    \text{BF}_{10} = 
    \begin{cases} 
      \frac{1}{-e \cdot p \ln p}, & p \le 1/e \\
      1, & p > 1/e 
    \end{cases}
    \label{eq:bayes_factor}
\end{equation}
\vspace{-5pt}

Finally, we convert this Bayes Factor into the continuous Contamination Confidence score $\mathcal{C}_{cont}$, a bound-based contamination evidence score. To avoid injecting subjective bias, we assume a neutral prior probability $\pi = P(H_1) = 0.5$. The final confidence score is calculated as follows, with the detailed derivation provided in Appendix~\ref{appendix:bayes_derivation}:
\begin{equation}
    \mathcal{C}_{cont} = \frac{\text{BF}_{10} \cdot \pi}{\text{BF}_{10} \cdot \pi + (1 - \pi)} = \frac{\text{BF}_{10}}{\text{BF}_{10} + 1}
    \label{eq:confidence_score}
\end{equation}
\vspace{-10pt}

If the performance gap is statistically insignificant ($p \ge 1/e \approx 0.368$), Equation~\ref{eq:bayes_factor} yields $\text{BF}_{10} = 1$, which subsequently results in $\mathcal{C}_{cont} = 0.5$ via Equation~\ref{eq:confidence_score}, correctly indicating no statistical evidence of contamination. Conversely, as the performance gap becomes highly significant ($p \to 0$), the Bayes Factor  ($\text{BF}_{10} \to \infty$). Consequently, the contamination confidence $\mathcal{C}_{cont}$ asymptotically approaches its ceiling of $1.0$, indicating overwhelming statistical evidence of contamination.


\vspace{-10pt}
\section{Experiments}
\label{sec:experiments}
\vspace{-5pt}

We comprehensively evaluate ZCP across four dimensions: (1) ``flipped experiments'' on existing models to validate our core approach (Section~\ref{sec:experiments-on-existing-models}); (2) controlled simulations of stealthy and evasive contamination via explicitly fine-tuned models (Section~\ref{sec:experiments-on-finetuned-models}); (3) detailed ablations on reasoning confounders and dataset scaling (Appendix~\ref{sec:further-analysis}); and (4) real-world auditing of state-of-the-art open-weight and closed-source commercial models (Appendix~\ref{sec:real-world}).

\vspace{-5pt}
\subsection{Experiments on Existing Models}
\label{sec:experiments-on-existing-models}
\vspace{-2pt}

To validate our method without prohibitive training costs, we first evaluate highly optimized existing models \citep{qwen2-5-math,GRPO}. Specifically, we simulate evasive data contamination using a ``flipped experiment'' paradigm, as detailed below.

\textbf{Models and Datasets.} We utilize Qwen2.5-Math-7B-Instruct and DeepSeek-Math-7B-RL as our target models. For contaminated data, we select the training splits of GSM8K \citep{gsm8k} and MATH \citep{math-dataset}, as their inclusion in the training data of the models is explicitly confirmed in the corresponding technical reports \citep{qwen2-5-math,GRPO}. Conversely, GSM1K \citep{gsm1k} serves as our strictly clean (uncontaminated) benchmark, since its publication postdates the training cutoffs of both models. To manage scale, all evaluations are conducted on representative random subsets (detailed statistics in Appendix~\ref{appendix:dataset_details}).

\textbf{Simulating Evasive Contamination.} To evaluate our method's robustness against evasive paraphrasing, we conduct a symmetric ``flipped experiment''. In the wild, models train on paraphrased data to excel on original benchmarks. Symmetrically, we evaluate target models (trained on original data) on aggressively paraphrased benchmark variants. We employ \texttt{gpt-4o} to rewrite the textual context while strictly preserving all original numerical values and logic (see Appendix~\ref{appendix:system_prompts} for prompts).

\begin{table*}[ht!]
    \centering
    \caption{Detection results of ZCP on existing reasoning models evaluated on contaminated benchmarks. For each metric, we report the value of the metric on the reference dataset ($S_{ref}$), the value of metric ($S$) and the Contamination Confidence ($\mathcal{C}_{cont}$) on both the original and paraphrased test variants, as mean $\pm$ standard deviation over 3 random seeds; $S_{ref}$ is pooled across the original and paraphrased runs (n=6). A confidence score of $\mathcal{C}_{cont} \to 1$ indicates definitive data contamination. The smallest effective $p$-value from the bootstrap test is $1.0e-4$, corresponding to $\mathcal{C}_{cont} \approx 0.998$. When the bootstrap $p$-value rounds to 0 in all three seeds, we denote $\mathcal{C}_{cont}$ as $>0.998$.}
    \label{tab:zcp_existing_models}
    \renewcommand{\arraystretch}{1.0}
    \setlength{\tabcolsep}{3pt}
    \resizebox{\textwidth}{!}{
    \begin{tabular}{lll c cc cc}
        \toprule
        \multirow{2}{*}{\textbf{Model}} & \multirow{2}{*}{\textbf{Data}} & \multirow{2}{*}{\textbf{Metric}} & \multirow{2}{*}{\textbf{$S_{ref}$}} & \multicolumn{2}{c}{\textbf{Original}} & \multicolumn{2}{c}{\textbf{Paraphrased}} \\
        \cmidrule(lr){5-6} \cmidrule(lr){7-8}
        & & & & \textbf{$S$} & \textbf{$\mathcal{C}_{cont}$} & \textbf{$S$} & \textbf{$\mathcal{C}_{cont}$} \\
        \midrule
        
        \multirow{8}{*}{\textbf{DeepSeek-Math}} 
        & \multirow{4}{*}{GSM8K} 
        &$ ACC (\%) $& 21.70 $\pm$ 0.33 & 29.73 $\pm$ 0.12 & 0.993 $\pm$ 0.006 & 27.33 $\pm$ 0.12 & 0.917 $\pm$ 0.016 \\
        & &$ Con (\%) $& 21.30 $\pm$ 0.37 & 29.93 $\pm$ 0.12 & 0.998 $\pm$ 0.003 & 26.53 $\pm$ 0.12 & 0.879 $\pm$ 0.030 \\
        & & $\mathcal{P}_{first}$ & 0.380 $\pm$ 0.000 & 0.463 $\pm$ 0.000 & $>$0.998 & 0.459 $\pm$ 0.000 & $>$0.998 \\
        & & $\mathcal{P}_{all}$ & 0.285 $\pm$ 0.000 & 0.395 $\pm$ 0.000 & $>$0.998 & 0.381 $\pm$ 0.000 & $>$0.998 \\
        \cmidrule{2-8}
        & \multirow{4}{*}{MATH} 
        &$ ACC (\%) $& 20.43 $\pm$ 0.00 & 37.29 $\pm$ 0.00 & 1.000 $\pm$ 0.000 & 30.00 $\pm$ 0.14 & 1.000 $\pm$ 0.000 \\
        & &$ Con (\%) $& 18.81 $\pm$ 0.53 & 31.29 $\pm$ 0.76 & 1.000 $\pm$ 0.000 & 25.76 $\pm$ 0.44 & 0.993 $\pm$ 0.005 \\
        & & $\mathcal{P}_{first}$ & 0.251 $\pm$ 0.000 & 0.403 $\pm$ 0.000 & $>$0.998 & 0.327 $\pm$ 0.000 & $>$0.998 \\
        & & $\mathcal{P}_{all}$ & 0.185 $\pm$ 0.000 & 0.347 $\pm$ 0.000 & $>$0.998 & 0.285 $\pm$ 0.000 & $>$0.998 \\
        
        \midrule
        
        \multirow{8}{*}{\textbf{Qwen-Math}} 
        & \multirow{4}{*}{GSM8K} 
        &$ ACC (\%) $& 33.20 $\pm$ 0.28 & 45.73 $\pm$ 0.12 & 1.000 $\pm$ 0.000 & 42.33 $\pm$ 0.12 & 0.996 $\pm$ 0.002 \\
        & &$ Con (\%) $& 32.93 $\pm$ 0.39 & 45.87 $\pm$ 0.31 & 1.000 $\pm$ 0.000 & 42.13 $\pm$ 0.12 & 0.997 $\pm$ 0.002 \\
        & & $\mathcal{P}_{first}$ & 0.488 $\pm$ 0.000 & 0.532 $\pm$ 0.000 & 0.964 $\pm$ 0.002 & 0.532 $\pm$ 0.000 & 0.936 $\pm$ 0.006 \\
        & & $\mathcal{P}_{all}$ & 0.412 $\pm$ 0.000 & 0.511 $\pm$ 0.000 & $>$0.998 & 0.498 $\pm$ 0.000 & $>$0.998 \\
        \cmidrule{2-8}
        & \multirow{4}{*}{MATH} 
        &$ ACC (\%) $& 31.27 $\pm$ 0.11 & 49.29 $\pm$ 0.00 & 1.000 $\pm$ 0.000 & 43.14 $\pm$ 0.00 & 1.000 $\pm$ 0.000 \\
        & &$ Con (\%) $& 30.09 $\pm$ 0.11 & 46.24 $\pm$ 0.30 & 1.000 $\pm$ 0.000 & 40.90 $\pm$ 0.50 & 1.000 $\pm$ 0.000 \\
        & & $\mathcal{P}_{first}$ & 0.304 $\pm$ 0.000 & 0.427 $\pm$ 0.000 & $>$0.998 & 0.400 $\pm$ 0.000 & $>$0.998 \\
        & & $\mathcal{P}_{all}$ & 0.277 $\pm$ 0.000 & 0.426 $\pm$ 0.000 & $>$0.998 & 0.388 $\pm$ 0.000 & $>$0.998 \\
        
        \bottomrule
    \end{tabular}
    }
    \vspace{-10pt}
\end{table*}

\vspace{-5pt}
\subsubsection{Results and Analysis}
\vspace{-2pt}

The results of our method are presented in Table~\ref{tab:zcp_existing_models}. ZCP perfectly unmasks both direct and evasive data contamination. All models exhibit a massive performance drop when moving from the original/paraphrased data to the reference dataset ($\tilde{D}_{eval}$) under the zero-CoT setting. We also translate this degradation into Contamination Confidence scores $\mathcal{C}_{cont}$ through our statistical framework, which approximate 1.000 across all four metrics on contaminated datasets GSM8K and MATH, even on the paraphrased datasets. 
Crucially, ZCP succeeds against evasive data contamination because it transcends surface-level verbatim matching. Instead, it targets the latent \texttt{question $\rightarrow$ answer} shortcut mapping that models internalize during contaminated training. By enforcing zero-CoT generation, ZCP directly triggers this shortcut mapping, which easily survives textual paraphrasing ($D'_{eval}$) but is fundamentally broken by the isomorphic numerical perturbations in our reference data $\tilde{D}_{eval}$. Consequently, contaminated models exhibit a severe performance drop on $\tilde{D}_{eval}$, thus yielding a high Contamination Confidence ($\mathcal{C}_{cont}$).


We can conduct experiments on the clean dataset GSM1K, whose publication date is later than these two models, the zero-CoT performance on the original GSM1K questions is statistically indistinguishable from the performance on the reference data $\tilde{D}_{eval}$, and the Contamination Confidence $\mathcal{C}_{cont} \approx 0.500$, as shown in Table \ref{tab:zcp_gsm1k}. This confirms that ZCP is strictly sensitive to data contamination and highly reliable against false positives in real-world auditing scenarios. Results for the OLMo family of open-data models, evaluated under the same flipped-experiment paradigm, are provided in Appendix~\ref{appendix:olmo}.

\begin{table}[htbp!]
    \centering
    \caption{Detection results of ZCP on the uncontaminated GSM1K benchmark, as mean $\pm$ standard deviation over 3 random seeds. The Contamination Confidence ($\mathcal{C}_{cont}$) remains near 0.500, indicating no statistical evidence of memorization.}
    \label{tab:zcp_gsm1k}
    \renewcommand{\arraystretch}{1.0}
    \resizebox{0.95\columnwidth}{!}{
    \begin{tabular}{lllc cc}
        \toprule
        \multirow{2}{*}{\textbf{Model}} & \multirow{2}{*}{\textbf{Data}} & \multirow{2}{*}{\textbf{Metric}} & \multirow{2}{*}{\textbf{$S_{ref}$}} & \multicolumn{2}{c}{\textbf{Original}} \\
        \cmidrule(lr){5-6}
        & & & & \textbf{$S$} & \textbf{$\mathcal{C}_{cont}$} \\
        \midrule
        
        \multirow{4}{*}{\textbf{DeepSeek-Math}} 
        & \multirow{4}{*}{GSM1K} 
        &$ ACC (\%) $& 18.33 $\pm$ 0.29 & 16.00 $\pm$ 0.00 & 0.500 $\pm$ 0.000 \\
        & &$ Con (\%) $& 16.50 $\pm$ 0.00 & 15.00 $\pm$ 0.00 & 0.500 $\pm$ 0.000 \\
        & & $\mathcal{P}_{first}$ & 0.331 $\pm$ 0.000 & 0.348 $\pm$ 0.000 & 0.509 $\pm$ 0.001 \\
        & & $\mathcal{P}_{all}$ & 0.233 $\pm$ 0.000 & 0.240 $\pm$ 0.000 & 0.500 $\pm$ 0.000 \\
        
        \midrule
        
        \multirow{4}{*}{\textbf{Qwen-Math}} 
        & \multirow{4}{*}{GSM1K} 
        &$ ACC (\%) $& 22.17 $\pm$ 0.29 & 23.33 $\pm$ 0.29 & 0.500 $\pm$ 0.000 \\
        & &$ Con (\%) $& 22.17 $\pm$ 0.29 & 22.83 $\pm$ 0.29 & 0.500 $\pm$ 0.000 \\
        & & $\mathcal{P}_{first}$ & 0.387 $\pm$ 0.000 & 0.410 $\pm$ 0.000 & 0.560 $\pm$ 0.001 \\
        & & $\mathcal{P}_{all}$ & 0.298 $\pm$ 0.000 & 0.316 $\pm$ 0.000 & 0.533 $\pm$ 0.000 \\
        
        \bottomrule
    \end{tabular}
    }
\end{table}

\vspace{-5pt}
\subsection{Experiments on Finetuned Models}
\label{sec:experiments-on-finetuned-models}
\vspace{-2pt}


Having validated ZCP on existing models through flipped experiments, we now escalate our evaluation to an authentic evasive data contamination setting. In this section, we actively finetune LLMs on paraphrased datasets and test ZCP on them.

\textbf{Models and Data.}
We evaluate two target models: Qwen2.5-Math-7B-Instruct, fine-tuned on the Omni-MATH benchmark \citep{omnimath}, and Qwen3-8B (non-thinking), fine-tuned on a multi-domain mixture (spanning physics, chemistry, business, and finance) from MMLU-Pro \citep{mmlu-pro} and XFINBENCH \citep{xfinbench}. Each dataset is evenly partitioned into a contaminated set (Dataset C) and a strictly held-out uncontaminated control (Dataset U). To simulate evasive contamination, we paraphrase Dataset C into six distinct variants for training, synthesizing reasoning chains for instances lacking them to ensure complete problem-reasoning-answer triplets. Finally, the resulting evasively contaminated models are evaluated directly on the original benchmarks.


\textbf{Training Pipeline.} We simulate the contamination process via LoRA fine-tuning using a standard two-stage paradigm to mirror modern state-of-the-art training paradigms. First, Supervised Fine-Tuning (SFT) trains the model to generate basic reasoning formats and final answers. Subsequently, Reinforcement Learning (RL) via GRPO further optimizes and incentivizes these reasoning capabilities. Comprehensive training details are provided in Appendix~\ref{appendix:training_details}.


\subsubsection{Effect of Evasive Data Contamination}

\begin{wraptable}{r}{0.5\textwidth} 
    \centering
    \vspace{-20pt}
    \small 
    \setlength{\tabcolsep}{4pt} 
    \caption{Accuracy (\%) before and after evasive data contamination on Dataset C and Dataset U of Omni-MATH (on Qwen2.5-Math) and Multi-domain Data (on Qwen3-8B).}
    \label{tab:effect-of-data-contamination}
    \begin{tabular}{@{} l cccc @{}}
        \toprule
        \multirow{2}{*}{\textbf{Dataset}} & \multicolumn{2}{c}{\textbf{Dataset C}} & \multicolumn{2}{c}{\textbf{Dataset U}} \\
        \cmidrule(lr){2-3} \cmidrule(lr){4-5}
        & Before & After & Before & After \\
        \midrule
        Omni-MATH         & 21.28 & 43.38 & 23.64 & 26.77 \\
        Multi-domain Data & 36.67 & 66.03 & 36.83 & 36.30 \\
        \bottomrule
    \end{tabular}
\end{wraptable}
The effect of our evasive contamination pipeline is detailed in Table~\ref{tab:effect-of-data-contamination}. We observe significant performance gains on Dataset C, whose paraphrased variants were exposed during training. Importantly, performance on the held-out Dataset U remains stable before and after fine-tuning. This contrast confirms that the improvements on Dataset C stem from data contamination, rather than a generalized enhancement in reasoning capabilities.

\subsubsection{Detection Results and Analysis}



\textbf{Results of ZCP.}
Our ZCP framework successfully detects this evasive data contamination. As presented in Table~\ref{tab:zcp_finetuned_combined}, ZCP yields high Contamination Confidence ($\mathcal{C}_{cont} \to 1.000$) across all performance metrics on Dataset C for both the finetuned Qwen-MATH and Qwen3 models. Furthermore, ZCP reliably outputs low Contamination Confidence ($\mathcal{C}_{cont} \approx 0.500$) on the uncontaminated Dataset U, demonstrating its robustness against false positives. These results on custom finetuned models definitively reinforce the effectiveness and reliability of ZCP in detecting evasive data contamination.

\begin{table*}[htbp]
    \centering
    \vspace{-10pt}
    \caption{Detection results of ZCP on finetuned (FT) models evaluated on \textbf{Dataset C} and \textbf{Dataset U} (Qwen-Math on Omni-MATH; Qwen3 on Multi-domain data), as mean $\pm$ standard deviation over 3 random seeds. The robust contrast between the high Contamination Confidence on Dataset C ($\mathcal{C}_{cont} \to 1.000$) and the low Contamination Confidence on Dataset U ($\mathcal{C}_{cont} \approx 0.500$) demonstrates the precision of ZCP.}
    \label{tab:zcp_finetuned_combined}
    \renewcommand{\arraystretch}{1.0}
    \setlength{\tabcolsep}{3pt}
    \resizebox{\textwidth}{!}{
    \begin{tabular}{lll ccc ccc}
        \toprule
        \multirow{3}{*}{\textbf{Model}} & \multirow{3}{*}{\textbf{Benchmark}} & \multirow{3}{*}{\textbf{Metric}} & \multicolumn{3}{c}{\textbf{Dataset C}} & \multicolumn{3}{c}{\textbf{Dataset U}} \\
        \cmidrule(lr){4-6} \cmidrule(lr){7-9}
        & & & \textbf{$S_{ref}$} & \textbf{$S$} & \textbf{$\mathcal{C}_{cont}$} & \textbf{$S_{ref}$} & \textbf{$S$} & \textbf{$\mathcal{C}_{cont}$} \\
        \midrule
        
        \multirow{4}{*}{\makecell{\textbf{FT Qwen-}\\\textbf{Math}}} 
        & \multirow{4}{*}{Omni-MATH} 
        &$ ACC (\%) $& 16.87 $\pm$ 0.03 & 24.26 $\pm$ 0.05 & 1.000 $\pm$ 0.000 & 12.30 $\pm$ 0.03 & 13.81 $\pm$ 0.05 & 0.551 $\pm$ 0.005 \\
        & &$ Con (\%) $& 21.83 $\pm$ 0.07 & 26.38 $\pm$ 0.20 & 1.000 $\pm$ 0.000 & 15.85 $\pm$ 0.07 & 17.13 $\pm$ 0.20 & 0.638 $\pm$ 0.043 \\
        & & $\mathcal{P}_{first}$ & 0.212 $\pm$ 0.000 & 0.334 $\pm$ 0.000 & $>$0.998 & 0.359 $\pm$ 0.000 & 0.375 $\pm$ 0.000 & 0.618 $\pm$ 0.000 \\
        & & $\mathcal{P}_{all}$ & 0.180 $\pm$ 0.000 & 0.305 $\pm$ 0.000 & $>$0.998 & 0.182 $\pm$ 0.000 & 0.194 $\pm$ 0.000 & 0.591 $\pm$ 0.000 \\
        
        \midrule
        
        \multirow{4}{*}{\textbf{FT Qwen3}} 
        & \multirow{4}{*}{\makecell{Multi-domain\\Data}} 
        &$ ACC (\%) $& 15.09 $\pm$ 0.39 & 25.01 $\pm$ 0.12 & 1.000 $\pm$ 0.000 & 14.42 $\pm$ 0.39 & 15.32 $\pm$ 0.12 & 0.567 $\pm$ 0.049 \\
        & &$ Con (\%) $& 18.99 $\pm$ 0.50 & 24.83 $\pm$ 0.49 & 1.000 $\pm$ 0.000 & 16.38 $\pm$ 0.50 & 17.21 $\pm$ 0.49 & 0.529 $\pm$ 0.033 \\
        & & $\mathcal{P}_{first}$ & 0.375 $\pm$ 0.000 & 0.470 $\pm$ 0.000 & $>$0.998 & 0.374 $\pm$ 0.000 & 0.375 $\pm$ 0.000 & 0.500 $\pm$ 0.000 \\
        & & $\mathcal{P}_{all}$ & 0.180 $\pm$ 0.000 & 0.297 $\pm$ 0.000 & $>$0.998 & 0.186 $\pm$ 0.000 & 0.193 $\pm$ 0.000 & 0.605 $\pm$ 0.000 \\
        
        \bottomrule
    \end{tabular}
    }
\end{table*}

\vspace{-10pt}
\section{Conclusion \& Limitation}
\label{sec:conclusion} 
\vspace{-5pt}

This paper addresses the critical threat of evasive data contamination in LLMs, where benchmarks are aggressively paraphrased to bypass traditional detection. We uncover a fundamental phenomenon: intermediate reasoning actively masks underlying memorization, acting as a severe confounder in contamination detection. 
Inspired by this, we introduce the Zero-CoT Probe (ZCP). By truncating reasoning chains and comparing zero-CoT performance against an isomorphically perturbed reference dataset, ZCP disentangles memorization from intrinsic ``superpower'', robustly exposing the latent shortcut mappings. We further propose \textit{Contamination Confidence}, a rigorous metric quantifying contamination severity, moving the community beyond brittle binary paradigms.
Our evaluations expose widespread contamination across prominent models, underscoring the necessity for transparent protocols.
For limitation, while ZCP ensures zero-CoT enforcement in open-weight models via direct token manipulation, extending this paradigm to closed-source APIs currently relies on careful prompt engineering. As commercial models become increasingly optimized for step-by-step reasoning, in the future, such prompts might not be as effective. We will leave this as an valuable future direction to further enhance closed-source models' data contamination auditing.

\subsection*{AI Use Statement}

In this work, we used generative AI tools for editing (e.g., grammar, spelling, and word choice) and for writing and debugging code used in our experimental pipeline and analysis scripts. We also used generative AI tools for data processing and filtering. We have not used generative AI tools for research ideation or experimental design. We have reviewed all AI-assisted work: AI-edited text was checked against the intended meaning by the authors, and AI-assisted code was tested and verified for correctness against expected outputs before being used to produce any reported result. We take responsibility for the final content of this work, including text, claims, or artifacts produced with the aid of generative AI.


\bibliographystyle{iclr2027_conference}
\bibliography{custom}

\begin{thebibliography}{55}
\providecommand{\natexlab}[1]{#1}
\providecommand{\url}[1]{\texttt{#1}}
\expandafter\ifx\csname urlstyle\endcsname\relax
  \providecommand{\doi}[1]{doi: #1}\else
  \providecommand{\doi}{doi: \begingroup \urlstyle{rm}\Url}\fi

\bibitem[Achiam et~al.(2023)Achiam, Adler, Agarwal, Ahmad, Akkaya, Aleman,
  Almeida, Altenschmidt, Altman, Anadkat, et~al.]{gpt4}
Josh Achiam, Steven Adler, Sandhini Agarwal, Lama Ahmad, Ilge Akkaya,
  Florencia~Leoni Aleman, Diogo Almeida, Janko Altenschmidt, Sam Altman,
  Shyamal Anadkat, et~al.
\newblock Gpt-4 technical report.
\newblock \emph{arXiv preprint arXiv:2303.08774}, 2023.

\bibitem[Bayarri et~al.(2016)Bayarri, Benjamin, Berger, and
  Sellke]{bayes-factor}
MJ~Bayarri, Daniel~J Benjamin, James~O Berger, and Thomas~M Sellke.
\newblock Rejection odds and rejection ratios: A proposal for statistical
  practice in testing hypotheses.
\newblock \emph{Journal of Mathematical Psychology}, 72:\penalty0 90--103,
  2016.

\bibitem[Brown et~al.(2020)Brown, Mann, Ryder, Subbiah, Kaplan, Dhariwal,
  Neelakantan, Shyam, Sastry, Askell, Agarwal, Herbert-Voss, Krueger, Henighan,
  Child, Ramesh, Ziegler, Wu, Winter, Hesse, Chen, Sigler, Litwin, Gray, Chess,
  Clark, Berner, McCandlish, Radford, Sutskever, and Amodei]{gpt3}
Tom~B. Brown, Benjamin Mann, Nick Ryder, Melanie Subbiah, Jared Kaplan,
  Prafulla Dhariwal, Arvind Neelakantan, Pranav Shyam, Girish Sastry, Amanda
  Askell, Sandhini Agarwal, Ariel Herbert-Voss, Gretchen Krueger, Tom Henighan,
  Rewon Child, Aditya Ramesh, Daniel~M. Ziegler, Jeffrey Wu, Clemens Winter,
  Christopher Hesse, Mark Chen, Eric Sigler, Mateusz Litwin, Scott Gray,
  Benjamin Chess, Jack Clark, Christopher Berner, Sam McCandlish, Alec Radford,
  Ilya Sutskever, and Dario Amodei.
\newblock Language models are few-shot learners.
\newblock In \emph{Proceedings of the 34th International Conference on Neural
  Information Processing Systems}, NIPS '20, Red Hook, NY, USA, 2020. Curran
  Associates Inc.
\newblock ISBN 9781713829546.

\bibitem[Carlini et~al.(2021)Carlini, Tramer, Wallace, Jagielski, Herbert-Voss,
  Lee, Roberts, Brown, Song, Erlingsson, et~al.]{perplexity}
Nicholas Carlini, Florian Tramer, Eric Wallace, Matthew Jagielski, Ariel
  Herbert-Voss, Katherine Lee, Adam Roberts, Tom Brown, Dawn Song, Ulfar
  Erlingsson, et~al.
\newblock Extracting training data from large language models.
\newblock In \emph{30th USENIX security symposium (USENIX Security 21)}, pp.\
  2633--2650, 2021.

\bibitem[Carlini et~al.(2023)Carlini, Ippolito, Jagielski, Lee, Tramer, and
  Zhang]{data-completion1}
Nicholas Carlini, Daphne Ippolito, Matthew Jagielski, Katherine Lee, Florian
  Tramer, and Chiyuan Zhang.
\newblock Quantifying memorization across neural language models.
\newblock In \emph{The Eleventh International Conference on Learning
  Representations}, 2023.
\newblock URL \url{https://openreview.net/forum?id=TatRHT_1cK}.

\bibitem[Cheng et~al.(2025)Cheng, Chang, and Wu]{survey2025}
Yuxing Cheng, Yi~Chang, and Yuan Wu.
\newblock A survey on data contamination for large language models.
\newblock \emph{arXiv preprint arXiv:2502.14425}, 2025.

\bibitem[Cobbe et~al.(2021)Cobbe, Kosaraju, Bavarian, Chen, Jun, Kaiser,
  Plappert, Tworek, Hilton, Nakano, et~al.]{gsm8k}
Karl Cobbe, Vineet Kosaraju, Mohammad Bavarian, Mark Chen, Heewoo Jun, Lukasz
  Kaiser, Matthias Plappert, Jerry Tworek, Jacob Hilton, Reiichiro Nakano,
  et~al.
\newblock Training verifiers to solve math word problems.
\newblock \emph{arXiv preprint arXiv:2110.14168}, 2021.

\bibitem[Comanici et~al.(2025)Comanici, Bieber, Schaekermann, Pasupat,
  Sachdeva, Dhillon, Blistein, Ram, Zhang, Rosen, et~al.]{gemini2-5}
Gheorghe Comanici, Eric Bieber, Mike Schaekermann, Ice Pasupat, Noveen
  Sachdeva, Inderjit Dhillon, Marcel Blistein, Ori Ram, Dan Zhang, Evan Rosen,
  et~al.
\newblock Gemini 2.5: Pushing the frontier with advanced reasoning,
  multimodality, long context, and next generation agentic capabilities.
\newblock \emph{arXiv preprint arXiv:2507.06261}, 2025.

\bibitem[Dekoninck et~al.(2024)Dekoninck, M{\"u}ller, Baader, Fischer, and
  Vechev]{evasive-data-contamination}
Jasper Dekoninck, Mark~Niklas M{\"u}ller, Maximilian Baader, Marc Fischer, and
  Martin Vechev.
\newblock Evading data contamination detection for language models is (too)
  easy.
\newblock \emph{arXiv preprint arXiv:2402.02823}, 2024.

\bibitem[Deng et~al.(2024)Deng, Zhao, Tang, Gerstein, and
  Cohan]{investigating-data-contamination}
Chunyuan Deng, Yilun Zhao, Xiangru Tang, Mark Gerstein, and Arman Cohan.
\newblock Investigating data contamination in modern benchmarks for large
  language models.
\newblock In \emph{Proceedings of the 2024 Conference of the North American
  Chapter of the Association for Computational Linguistics: Human Language
  Technologies (Volume 1: Long Papers)}, pp.\  8706--8719, 2024.

\bibitem[Dong et~al.(2024)Dong, Jiang, Liu, Jin, Gu, Yang, and Li]{ccd-method}
Yihong Dong, Xue Jiang, Huanyu Liu, Zhi Jin, Bin Gu, Mengfei Yang, and Ge~Li.
\newblock Generalization or memorization: Data contamination and trustworthy
  evaluation for large language models.
\newblock \emph{arXiv preprint arXiv:2402.15938}, 2024.

\bibitem[Efron(1992)]{bootstrap-test}
Bradley Efron.
\newblock Bootstrap methods: another look at the jackknife.
\newblock In \emph{Breakthroughs in statistics: Methodology and distribution},
  pp.\  569--593. Springer, 1992.

\bibitem[Elangovan et~al.(2021)Elangovan, He, and
  Verspoor]{quantify-data-leakage}
Aparna Elangovan, Jiayuan He, and Karin Verspoor.
\newblock Memorization vs. generalization: Quantifying data leakage in nlp
  performance evaluation.
\newblock In \emph{Proceedings of the 16th Conference of the European Chapter
  of the Association for Computational Linguistics: Main Volume}, pp.\
  1325--1335, 2021.

\bibitem[Fu et~al.(2025)Fu, Uzuner, Yetisgen-Yildiz, and Xia]{survey3}
Yujuan Fu, Ozlem Uzuner, Meliha Yetisgen-Yildiz, and Fei Xia.
\newblock Does data contamination detection work (well) for llms? a survey and
  evaluation on detection assumptions.
\newblock In \emph{Findings of the Association for Computational Linguistics:
  NAACL 2025}, pp.\  5235--5256, 2025.

\bibitem[Gao et~al.(2025)Gao, Song, Yang, Cai, Miao, Dong, Li, Ma, Chen, Xu,
  Tang, Wang, Zan, Quan, Zhang, Sha, Zhang, Ren, Liu, and Chang]{omnimath}
Bofei Gao, Feifan Song, Zhe Yang, Zefan Cai, Yibo Miao, Qingxiu Dong, Lei Li,
  Chenghao Ma, Liang Chen, Runxin Xu, Zhengyang Tang, Benyou Wang, Daoguang
  Zan, Shanghaoran Quan, Ge~Zhang, Lei Sha, Yichang Zhang, Xuancheng Ren,
  Tianyu Liu, and Baobao Chang.
\newblock Omni-{MATH}: A universal olympiad level mathematic benchmark for
  large language models.
\newblock In \emph{The Thirteenth International Conference on Learning
  Representations}, 2025.
\newblock URL \url{https://openreview.net/forum?id=yaqPf0KAlN}.

\bibitem[Golchin \& Surdeanu(2023)Golchin and Surdeanu]{data-reconstruction}
Shahriar Golchin and Mihai Surdeanu.
\newblock Time travel in llms: Tracing data contamination in large language
  models.
\newblock \emph{arXiv preprint arXiv:2308.08493}, 2023.

\bibitem[Grattafiori et~al.(2024)Grattafiori, Dubey, Jauhri, Pandey, Kadian,
  Al-Dahle, Letman, Mathur, Schelten, Vaughan, et~al.]{llama3}
Aaron Grattafiori, Abhimanyu Dubey, Abhinav Jauhri, Abhinav Pandey, Abhishek
  Kadian, Ahmad Al-Dahle, Aiesha Letman, Akhil Mathur, Alan Schelten, Alex
  Vaughan, et~al.
\newblock The llama 3 herd of models.
\newblock \emph{arXiv preprint arXiv:2407.21783}, 2024.

\bibitem[Hendrycks et~al.(2021{\natexlab{a}})Hendrycks, Burns, Basart, Zou,
  Mazeika, Song, and Steinhardt]{mmlu}
Dan Hendrycks, Collin Burns, Steven Basart, Andy Zou, Mantas Mazeika, Dawn
  Song, and Jacob Steinhardt.
\newblock Measuring massive multitask language understanding.
\newblock In \emph{International Conference on Learning Representations},
  2021{\natexlab{a}}.
\newblock URL \url{https://openreview.net/forum?id=d7KBjmI3GmQ}.

\bibitem[Hendrycks et~al.(2021{\natexlab{b}})Hendrycks, Burns, Kadavath, Arora,
  Basart, Tang, Song, and Steinhardt]{math-dataset}
Dan Hendrycks, Collin Burns, Saurav Kadavath, Akul Arora, Steven Basart, Eric
  Tang, Dawn Song, and Jacob Steinhardt.
\newblock Measuring mathematical problem solving with the math dataset.
\newblock \emph{arXiv preprint arXiv:2103.03874}, 2021{\natexlab{b}}.

\bibitem[Huang et~al.(2025)Huang, Guo, Li, Ji, Ge, Li, Guo, Cai, Yuan, Wang,
  Wu, Yin, Tang, Huang, Jin, Chen, Zhang, and Wang]{math-perturb}
Kaixuan Huang, Jiacheng Guo, Zihao Li, Xiang Ji, Jiawei Ge, Wenzhe Li, Yingqing
  Guo, Tianle Cai, Hui Yuan, Runzhe Wang, Yue Wu, Ming Yin, Shange Tang,
  Yangsibo Huang, Chi Jin, Xinyun Chen, Chiyuan Zhang, and Mengdi Wang.
\newblock Math-perturb: Benchmarking llms' math reasoning abilities against
  hard perturbations.
\newblock \emph{arXiv preprint arXiv:2502.06453}, 2025.

\bibitem[Ippolito et~al.(2023)Ippolito, Tramer, Nasr, Zhang, Jagielski, Lee,
  Choo, and Carlini]{critic-verbatim-memorization}
Daphne Ippolito, Florian Tramer, Milad Nasr, Chiyuan Zhang, Matthew Jagielski,
  Katherine Lee, Christopher~Choquette Choo, and Nicholas Carlini.
\newblock Preventing generation of verbatim memorization in language models
  gives a false sense of privacy.
\newblock In \emph{Proceedings of the 16th International Natural Language
  Generation Conference}, pp.\  28--53, 2023.

\bibitem[Jaech et~al.(2024)Jaech, Kalai, Lerer, Richardson, El-Kishky, Low,
  Helyar, Madry, Beutel, Carney, et~al.]{o1}
Aaron Jaech, Adam Kalai, Adam Lerer, Adam Richardson, Ahmed El-Kishky, Aiden
  Low, Alec Helyar, Aleksander Madry, Alex Beutel, Alex Carney, et~al.
\newblock Openai o1 system card.
\newblock \emph{arXiv preprint arXiv:2412.16720}, 2024.

\bibitem[Jimenez et~al.(2024)Jimenez, Yang, Wettig, Yao, Pei, Press, and
  Narasimhan]{swebench}
Carlos~E Jimenez, John Yang, Alexander Wettig, Shunyu Yao, Kexin Pei, Ofir
  Press, and Karthik~R Narasimhan.
\newblock {SWE}-bench: Can language models resolve real-world github issues?
\newblock In \emph{The Twelfth International Conference on Learning
  Representations}, 2024.
\newblock URL \url{https://openreview.net/forum?id=VTF8yNQM66}.

\bibitem[Kumar et~al.(2022)Kumar, Maheshwary, and Pudi]{kumar2022practice}
Vivek Kumar, Rishabh Maheshwary, and Vikram Pudi.
\newblock Practice makes a solver perfect: Data augmentation for math word
  problem solvers.
\newblock In \emph{Proceedings of the 2022 Conference of the North American
  Chapter of the Association for Computational Linguistics: Human Language
  Technologies}, 2022.

\bibitem[Lanham et~al.(2023)Lanham, Chen, Radhakrishnan, Steiner, Denison,
  Hernandez, Li, Durmus, Hubinger, Kernion, et~al.]{faithfulness-of-cot}
Tamera Lanham, Anna Chen, Ansh Radhakrishnan, Benoit Steiner, Carson Denison,
  Danny Hernandez, Dustin Li, Esin Durmus, Evan Hubinger, Jackson Kernion,
  et~al.
\newblock Measuring faithfulness in chain-of-thought reasoning.
\newblock \emph{arXiv preprint arXiv:2307.13702}, 2023.

\bibitem[Li et~al.(2024)Li, Cui, Zhao, Kong, and Bi]{gsm-plus}
Qintong Li, Leyang Cui, Xueliang Zhao, Lingpeng Kong, and Wei Bi.
\newblock Gsm-plus: A comprehensive benchmark for evaluating the robustness of
  llms as mathematical problem solvers.
\newblock \emph{arXiv preprint arXiv:2402.19255}, 2024.

\bibitem[Mattern et~al.(2023)Mattern, Mireshghallah, Jin, Sch{\"o}lkopf,
  Sachan, and Berg-Kirkpatrick]{member-inference}
Justus Mattern, Fatemehsadat Mireshghallah, Zhijing Jin, Bernhard
  Sch{\"o}lkopf, Mrinmaya Sachan, and Taylor Berg-Kirkpatrick.
\newblock Membership inference attacks against language models via
  neighbourhood comparison.
\newblock In \emph{Findings of the Association for Computational Linguistics:
  ACL 2023}, pp.\  11330--11343, 2023.

\bibitem[McNemar(1947)]{mcnemar-test}
Quinn McNemar.
\newblock Note on the sampling error of the difference between correlated
  proportions or percentages.
\newblock \emph{Psychometrika}, 12\penalty0 (2):\penalty0 153--157, 1947.

\bibitem[Mireshghallah et~al.(2022)Mireshghallah, Goyal, Uniyal,
  Berg-Kirkpatrick, and Shokri]{reference-model2}
Fatemehsadat Mireshghallah, Kartik Goyal, Archit Uniyal, Taylor
  Berg-Kirkpatrick, and Reza Shokri.
\newblock Quantifying privacy risks of masked language models using membership
  inference attacks.
\newblock In Yoav Goldberg, Zornitsa Kozareva, and Yue Zhang (eds.),
  \emph{Proceedings of the 2022 Conference on Empirical Methods in Natural
  Language Processing}, pp.\  8332--8347, Abu Dhabi, United Arab Emirates,
  December 2022. Association for Computational Linguistics.
\newblock \doi{10.18653/v1/2022.emnlp-main.570}.
\newblock URL \url{https://aclanthology.org/2022.emnlp-main.570/}.

\bibitem[Mirzadeh et~al.(2024)Mirzadeh, Alizadeh, Shahrokhi, Tuzel, Bengio, and
  Farajtabar]{gsm-symbolic}
Iman Mirzadeh, Keivan Alizadeh, Hooman Shahrokhi, Oncel Tuzel, Samy Bengio, and
  Mehrdad Farajtabar.
\newblock Gsm-symbolic: Understanding the limitations of mathematical reasoning
  in large language models.
\newblock \emph{arXiv preprint arXiv:2410.05229}, 2024.

\bibitem[{OLMo Team}(2025)]{olmo3}
{OLMo Team}.
\newblock Olmo 3: Charting a path through the model flow to lead open-source
  ai.
\newblock \emph{arXiv preprint arXiv:2512.13961}, 2025.

\bibitem[{OLMo Team} et~al.(2025){OLMo Team}, Walsh, Soldaini, Groeneveld,
  et~al.]{olmo2}
{OLMo Team}, Pete Walsh, Luca Soldaini, Dirk Groeneveld, et~al.
\newblock 2 olmo 2 furious.
\newblock \emph{arXiv preprint arXiv:2501.00656}, 2025.

\bibitem[Oren et~al.(2023)Oren, Meister, Chatterji, Ladhak, and
  Hashimoto]{shuffle-multiple-choice}
Yonatan Oren, Nicole Meister, Niladri~S Chatterji, Faisal Ladhak, and Tatsunori
  Hashimoto.
\newblock Proving test set contamination in black-box language models.
\newblock In \emph{The Twelfth International Conference on Learning
  Representations}, 2023.

\bibitem[Paul et~al.(2024)Paul, West, Bosselut, and
  Faltings]{faithfulness-of-cot2}
Debjit Paul, Robert West, Antoine Bosselut, and Boi Faltings.
\newblock Making reasoning matter: Measuring and improving faithfulness of
  chain-of-thought reasoning.
\newblock In \emph{Findings of the Association for Computational Linguistics:
  EMNLP 2024}, pp.\  15012--15032, 2024.

\bibitem[Rein et~al.(2024)Rein, Hou, Stickland, Petty, Pang, Dirani, Michael,
  and Bowman]{gpqa}
David Rein, Betty~Li Hou, Asa~Cooper Stickland, Jackson Petty, Richard~Yuanzhe
  Pang, Julien Dirani, Julian Michael, and Samuel~R. Bowman.
\newblock {GPQA}: A graduate-level google-proof q\&a benchmark.
\newblock In \emph{First Conference on Language Modeling}, 2024.
\newblock URL \url{https://openreview.net/forum?id=Ti67584b98}.

\bibitem[S{\'a}nchez~Salido et~al.(2025)S{\'a}nchez~Salido, Gonzalo, and
  Marco]{sanchezsalido2025none}
Eva S{\'a}nchez~Salido, Julio Gonzalo, and Guillermo Marco.
\newblock None of the others: a general technique to distinguish reasoning from
  memorization in multiple-choice llm evaluation benchmarks.
\newblock \emph{arXiv preprint arXiv:2502.12896}, 2025.

\bibitem[Schwarzschild et~al.(2025)Schwarzschild, Feng, Maini, Lipton, and
  Kolter]{data-completion2}
Avi Schwarzschild, Zhili Feng, Pratyush Maini, Zachary~Chase Lipton, and J~Zico
  Kolter.
\newblock Rethinking {LLM} memorization through the lens of adversarial
  compression.
\newblock In \emph{Red Teaming GenAI: What Can We Learn from Adversaries?},
  2025.
\newblock URL \url{https://openreview.net/forum?id=oMOoNzcuFO}.

\bibitem[Sellke et~al.(2001)Sellke, Bayarri, and Berger]{sellke-calibration}
Thomas Sellke, Mar{\'\i}a~Jes{\'u}s Bayarri, and James~O Berger.
\newblock Calibration of $\rho$ values for testing precise null hypotheses.
\newblock \emph{The American Statistician}, 55\penalty0 (1):\penalty0 62--71,
  2001.

\bibitem[Shao et~al.(2024)Shao, Wang, Zhu, Xu, Song, Bi, Zhang, Zhang, Li, Wu,
  et~al.]{GRPO}
Zhihong Shao, Peiyi Wang, Qihao Zhu, Runxin Xu, Junxiao Song, Xiao Bi, Haowei
  Zhang, Mingchuan Zhang, YK~Li, Yang Wu, et~al.
\newblock Deepseekmath: Pushing the limits of mathematical reasoning in open
  language models.
\newblock \emph{arXiv preprint arXiv:2402.03300}, 2024.

\bibitem[Shi(2023)]{DPCC}
Weijia Shi.
\newblock Detect-pretrain-code-contamination.
\newblock \url{https://github.com/swj0419/detect-pretrain-code-contamination},
  2023.

\bibitem[Shi et~al.(2023)Shi, Ajith, Xia, Huang, Liu, Blevins, Chen, and
  Zettlemoyer]{mink-prob}
Weijia Shi, Anirudh Ajith, Mengzhou Xia, Yangsibo Huang, Daogao Liu, Terra
  Blevins, Danqi Chen, and Luke Zettlemoyer.
\newblock Detecting pretraining data from large language models.
\newblock \emph{arXiv preprint arXiv:2310.16789}, 2023.

\bibitem[Student(1908)]{t-test}
Student.
\newblock The probable error of a mean.
\newblock \emph{Biometrika}, pp.\  1--25, 1908.

\bibitem[Tu et~al.(2024)Tu, Zhu, Bai, Yao, Hou, and Li]{reference-model3}
Shangqing Tu, Kejian Zhu, Yushi Bai, Zijun Yao, Lei Hou, and Juanzi Li.
\newblock Dice: Detecting in-distribution contamination in llm's fine-tuning
  phase for math reasoning.
\newblock \emph{arXiv preprint arXiv:2406.04197}, 2024.

\bibitem[Veselovsky et~al.(2023)Veselovsky, Ribeiro, and
  West]{indirect-data-contamination-by-llm}
Veniamin Veselovsky, Manoel~Horta Ribeiro, and Robert West.
\newblock Artificial artificial artificial intelligence: Crowd workers widely
  use large language models for text production tasks.
\newblock \emph{arXiv preprint arXiv:2306.07899}, 2023.

\bibitem[Wang et~al.(2026)Wang, Joshi, Plank, Angell, and He]{reward-hacking}
Xinpeng Wang, Nitish Joshi, Barbara Plank, Rico Angell, and He~He.
\newblock Is it thinking or cheating? detecting implicit reward hacking by
  measuring reasoning effort.
\newblock In \emph{The Fourteenth International Conference on Learning
  Representations}, 2026.
\newblock URL \url{https://openreview.net/forum?id=Gk7gLAtVDO}.

\bibitem[Wang et~al.(2024)Wang, Ma, Zhang, Ni, Chandra, Guo, Ren, Arulraj, He,
  Jiang, et~al.]{mmlu-pro}
Yubo Wang, Xueguang Ma, Ge~Zhang, Yuansheng Ni, Abhranil Chandra, Shiguang Guo,
  Weiming Ren, Aaran Arulraj, Xuan He, Ziyan Jiang, et~al.
\newblock Mmlu-pro: A more robust and challenging multi-task language
  understanding benchmark.
\newblock \emph{Advances in Neural Information Processing Systems},
  37:\penalty0 95266--95290, 2024.

\bibitem[Wei et~al.(2022)Wei, Wang, Schuurmans, Bosma, Xia, Chi, Le, Zhou,
  et~al.]{cot}
Jason Wei, Xuezhi Wang, Dale Schuurmans, Maarten Bosma, Fei Xia, Ed~Chi, Quoc~V
  Le, Denny Zhou, et~al.
\newblock Chain-of-thought prompting elicits reasoning in large language
  models.
\newblock \emph{Advances in neural information processing systems},
  35:\penalty0 24824--24837, 2022.

\bibitem[Wu et~al.(2025)Wu, Zhang, Dong, Xi, Zhao, Jin, Fan, Zhou, Lv, Zhang,
  et~al.]{qwen-math-data-contamination}
Mingqi Wu, Zhihao Zhang, Qiaole Dong, Zhiheng Xi, Jun Zhao, Senjie Jin, Xiaoran
  Fan, Yuhao Zhou, Huijie Lv, Ming Zhang, et~al.
\newblock Reasoning or memorization? unreliable results of reinforcement
  learning due to data contamination.
\newblock \emph{arXiv preprint arXiv:2507.10532}, 2025.

\bibitem[Xu et~al.(2024)Xu, Guan, Greene, Kechadi, et~al.]{2024survey1}
Cheng Xu, Shuhao Guan, Derek Greene, M~Kechadi, et~al.
\newblock Benchmark data contamination of large language models: A survey.
\newblock \emph{arXiv preprint arXiv:2406.04244}, 2024.

\bibitem[Yang et~al.(2024)Yang, Zhang, Hui, Gao, Yu, Li, Liu, Tu, Zhou, Lin,
  et~al.]{qwen2-5-math}
An~Yang, Beichen Zhang, Binyuan Hui, Bofei Gao, Bowen Yu, Chengpeng Li,
  Dayiheng Liu, Jianhong Tu, Jingren Zhou, Junyang Lin, et~al.
\newblock Qwen2. 5-math technical report: Toward mathematical expert model via
  self-improvement.
\newblock \emph{arXiv preprint arXiv:2409.12122}, 2024.

\bibitem[Yang et~al.(2025)Yang, Li, Yang, Zhang, Hui, Zheng, Yu, Gao, Huang,
  Lv, et~al.]{qwen3}
An~Yang, Anfeng Li, Baosong Yang, Beichen Zhang, Binyuan Hui, Bo~Zheng, Bowen
  Yu, Chang Gao, Chengen Huang, Chenxu Lv, et~al.
\newblock Qwen3 technical report.
\newblock \emph{arXiv preprint arXiv:2505.09388}, 2025.

\bibitem[Yang et~al.(2023)Yang, Chiang, Zheng, Gonzalez, and
  Stoica]{llm-decontaminator}
Shuo Yang, Wei-Lin Chiang, Lianmin Zheng, Joseph~E Gonzalez, and Ion Stoica.
\newblock Rethinking benchmark and contamination for language models with
  rephrased samples.
\newblock \emph{arXiv preprint arXiv:2311.04850}, 2023.

\bibitem[Zhang et~al.(2024)Zhang, Da, Lee, Robinson, Wu, Song, Zhao, Raja,
  Zhuang, Slack, et~al.]{gsm1k}
Hugh Zhang, Jeff Da, Dean Lee, Vaughn Robinson, Catherine Wu, William Song,
  Tiffany Zhao, Pranav Raja, Charlotte Zhuang, Dylan Slack, et~al.
\newblock A careful examination of large language model performance on grade
  school arithmetic.
\newblock \emph{Advances in Neural Information Processing Systems},
  37:\penalty0 46819--46836, 2024.

\bibitem[Zhang et~al.(2025)Zhang, Cao, and Liao]{xfinbench}
Zhihan Zhang, Yixin Cao, and Lizi Liao.
\newblock Xfinbench: Benchmarking llms in complex financial problem solving and
  reasoning.
\newblock In \emph{ACL (Findings)}, pp.\  8715--8758, 2025.
\newblock URL \url{https://aclanthology.org/2025.findings-acl.457/}.

\bibitem[Zhang et~al.(2022)Zhang, Zhang, Li, and Smola]{auto-cot}
Zhuosheng Zhang, Aston Zhang, Mu~Li, and Alex Smola.
\newblock Automatic chain of thought prompting in large language models.
\newblock \emph{arXiv preprint arXiv:2210.03493}, 2022.

\end{thebibliography}

\raggedbottom
\appendix

\section{Derivation of Contamination Confidence from Bayes Factor}
\label{appendix:bayes_derivation}

In Section~\ref{section:contamination-confidence}, we defined the Contamination Confidence $\mathcal{C}_{cont}$ as an upper bound on the posterior probability $P(H_1 \mid \text{data})$, obtained by substituting our upper-bound Bayes Factor ($\text{BF}_{10}$) into the standard posterior-odds transform. This appendix provides a concise derivation of this transform.

According to Bayes' theorem, the posterior odds of hypothesis $H_1$ (contaminated) versus $H_0$ (uncontaminated) given the observed data $D$ can be expressed as the product of the Bayes Factor and the prior odds:
\begin{equation}
    \underbrace{\frac{P(H_1 \mid D)}{P(H_0 \mid D)}}_{\text{Posterior Odds}} = \underbrace{\frac{P(D \mid H_1)}{P(D \mid H_0)}}_{\text{Bayes Factor } (\text{BF}_{10})} \cdot \underbrace{\frac{P(H_1)}{P(H_0)}}_{\text{Prior Odds}}
    \label{eq:odds_form}
\end{equation}

Let $\pi = P(H_1)$ denote the prior probability of contamination. Since hypotheses $H_1$ and $H_0$ are mutually exclusive and exhaustive, we have $P(H_0) = 1 - \pi$. Similarly, the posterior probabilities sum to one, meaning $P(H_0 \mid D) = 1 - P(H_1 \mid D)$. Substituting these terms into Equation~\ref{eq:odds_form} yields:
\begin{equation}
    \frac{P(H_1 \mid D)}{1 - P(H_1 \mid D)} = \text{BF}_{10} \cdot \frac{\pi}{1 - \pi}
    \label{eq:substitute}
\end{equation}

Rearranging Equation~\ref{eq:substitute} to isolate $P(H_1 \mid D)$ gives the exact posterior probability as a function of the true Bayes Factor. Since our $\text{BF}_{10}$ (Equation~\ref{eq:bayes_factor}) is only an upper bound on this true Bayes Factor, and the transform below is monotonically increasing in $\text{BF}_{10}$, substituting our upper-bound $\text{BF}_{10}$ yields the Contamination Confidence $\mathcal{C}_{cont}$ as an upper bound on the true posterior:
\begin{equation}
    \mathcal{C}_{cont} = \frac{\text{BF}_{10} \cdot \frac{\pi}{1 - \pi}}{1 + \text{BF}_{10} \cdot \frac{\pi}{1 - \pi}} = \frac{\text{BF}_{10} \cdot \pi}{\text{BF}_{10} \cdot \pi + (1 - \pi)} \ge P(H_1 \mid D)
    \label{eq:general_confidence_appendix}
\end{equation}

To prevent the injection of subjective bias into our detection metric, we strictly assume a neutral (uninformative) prior probability of $\pi = 0.5$. Under this assumption, the prior odds evaluate to $1$, naturally reducing Equation~\ref{eq:general_confidence_appendix} to our final applied formula:
\begin{equation}
    \mathcal{C}_{cont} = \frac{\text{BF}_{10} \cdot 0.5}{\text{BF}_{10} \cdot 0.5 + 0.5} = \frac{\text{BF}_{10}}{\text{BF}_{10} + 1}
\end{equation}

\section{Multi-model System for Reference Data Construction}
\label{appendix:multi-model-system}

\begin{figure}[tbhp]
    \centering
    \includegraphics[width=0.95\linewidth]{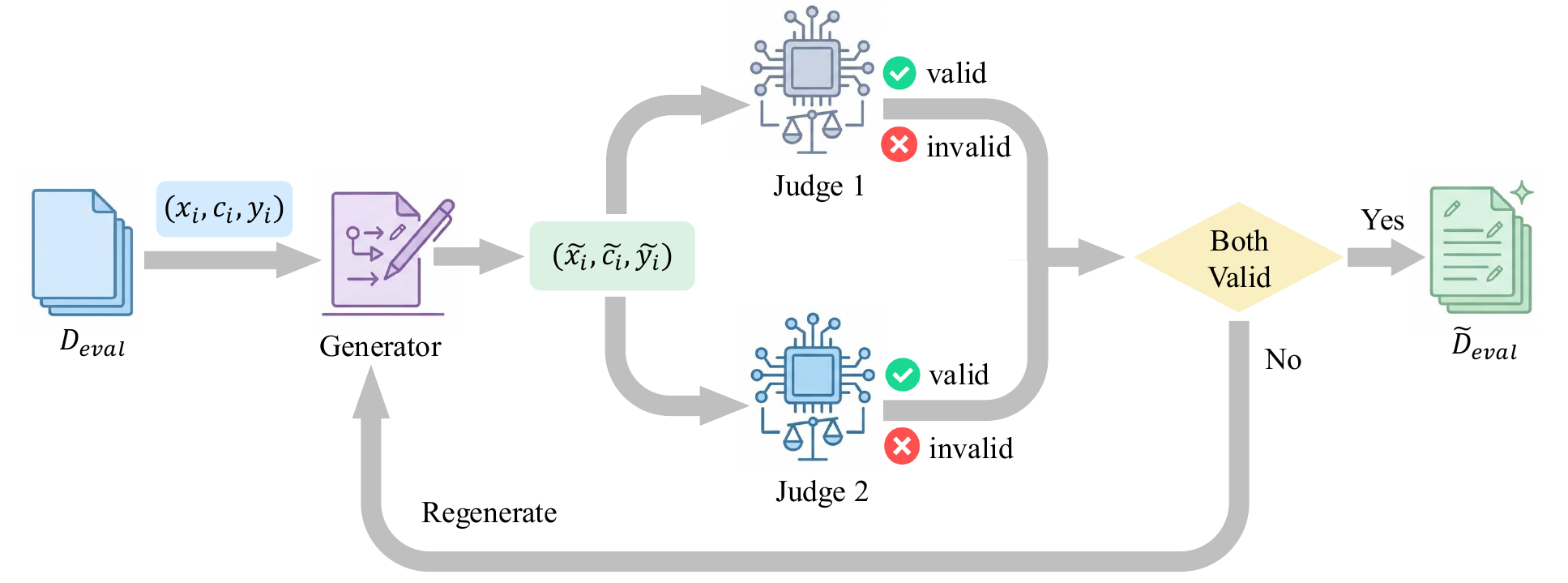}
    \caption{The automated multi-model pipeline for constructing the reference dataset $\tilde{D}_{eval}$. A generator creates isomorphically perturbed samples, which are incorporated into $\tilde{D}_{eval}$ only if two independent judge models reach a strict consensus on their validity.}
    \label{fig:multi-model}
\end{figure}

To execute our cleaning strategy in Section \ref{sec:reference} at scale, we design an automated, multi-model generation pipeline to synthesize the reference dataset $\tilde{D}_{eval}$, as illustrated in Figure \ref{fig:multi-model}. This system comprises a generator LLM and two independent judge LLMs. For each original triplet $(x_i, c_i, y_i) \in D_{eval}$, the generator first applies our isomorphic perturbation to the original question $x_i$, synthesizing the cleaned question $\tilde{x}_i$. Subsequently, it adapts the reasoning steps to form the new intermediate solution $\tilde{c}_i$, which naturally yields the recalculated final answer $\tilde{y}_i$. The judge models then rigorously verify the mathematical correctness of the generated triplet $(\tilde{x}_i, \tilde{c}_i, \tilde{y}_i)$. Synthesized samples are incorporated into $\tilde{D}_{eval}$ only if both judges reach a strict consensus regarding their validity; otherwise, the generator is prompted to regenerate the sample. The system prompts we use for each LLM are presented in Appendix~\ref{appendix:system_prompts}.

\section{Human Evaluation of Reference Data Quality}
\label{appendix:human-eval}

The multi-model pipeline in Appendix~\ref{appendix:multi-model-system} relies on LLM judges to verify the reference dataset $\tilde{D}_{eval}$. To further verify the quality of $\tilde{D}_{eval}$ with human judgment, we conducted a human study on the generated reference data for GSM8K and MATH. We invited 3 PhD students and 2 undergraduate students to evaluate correctness and difficulty; each rater judged 20 randomly sampled instances per benchmark (100 samples for GSM8K and MATH each). For difficulty, raters chose among ``original harder,'' ``reference harder,'' or ``equal.'' Results are in Table~\ref{tab:human_eval}: the reference data is highly accurate, and its difficulty is judged similar to the original in the large majority of cases. Since each sampled instance was judged by a single rater rather than by all five, we do not report inter-rater agreement.

\begin{table}[htbp]
    \centering
    \caption{Human evaluation of the reference dataset $\tilde{D}_{eval}$ on GSM8K and MATH (100 samples each).}
    \label{tab:human_eval}
    \renewcommand{\arraystretch}{1.0}
    \small
    \begin{tabular}{lcccc}
        \toprule
        \textbf{Benchmark} & \textbf{Accurate Samples (/100)} & \textbf{Original Harder} & \textbf{Reference Harder} & \textbf{Equal Difficulty} \\
        \midrule
        GSM8K & 99 & 2 & 5 & 93 \\
        MATH  & 99 & 2 & 4 & 94 \\
        \bottomrule
    \end{tabular}
\end{table}

\section{System Prompts for Data Construction}
\label{appendix:system_prompts}

In this appendix, we provide the exact system prompts utilized by our automated multi-model pipeline for constructing reference data $\tilde{D}_{eval}$. As discussed in Section~\ref{sec:experiments-on-existing-models} and Section~\ref{sec:experiments-on-finetuned-models}, our experiments require meticulously crafted data to either isolate memorization or simulate stealthy contamination. The pipeline relies on three distinct prompt templates:

\textbf{1. Isomorphic Perturbation Prompt (for Reference Cleaned Data $\tilde{D}_{eval}$):} 
Table~\ref{tab:prompt_cleaned_data} displays the prompt used to generate the reference dataset $\tilde{D}_{eval}$. This prompt strictly instructs the Generator LLM (we use \texttt{GPT-o3-mini}) to execute an isomorphic perturbation—altering the semantic narrative and numerical values while meticulously preserving the original order of magnitude, logical structure, and mathematical difficulty.

\textbf{2. Evasive Paraphrasing Prompt (for Paraphrased Data $D'_{eval}$):} 
Table~\ref{tab:prompt_evasive_data} presents the prompt used to synthesize the evasively contaminated data for finetuning our target models or for the clipped experiments. Unlike the perturbation prompt, this instruction forces the LLM to aggressively vary the textual syntax and entities while strictly retaining all original numerical values and the exact mathematical answer. 

\textbf{3. Mathematical Judge Prompt (for Consensus Verification):} 
Table~\ref{tab:prompt_judge} illustrates the prompt deployed to the two independent Judge LLMs (we use \texttt{GPT-o4-mini} and \texttt{Gemini-2.5-flash}). The model is tasked with rigorously verifying the mathematical equivalence and correctness of the generated problems, solutions, and final answers, ensuring the high quality of our constructed datasets.

\begin{table*}[htbp]
    \centering
    \small 
    \caption{The system prompt for generating the reference data ($\tilde{D}_{eval}$). It enforces isomorphic perturbation by modifying both the textual context and numerical values while preserving the mathematical logic.}
    \label{tab:prompt_cleaned_data}
    \renewcommand{\arraystretch}{1.0} 
    \begin{tabularx}{\textwidth}{@{} X @{}} 
        \toprule
        \multicolumn{1}{c}{\textbf{System Prompt: Reference Data Generator}} \\
        \midrule
        
        \textbf{Role \& Task:} \newline
        You are a mathematical data generator specialized in creating diverse training samples. Your task is to create a new sample by paraphrasing and modifying the original problem while maintaining the same difficulty level and solution logic. \\
        \midrule
        
        \textbf{1. Paraphrase \& Modify Problem:}
        \vspace{0.3em} 
        \begin{itemize}[leftmargin=1.5em, nosep]
            \item \textbf{Paraphrase}: Rephrase sentences, change wording, and adjust sentence structure to create a distinctly different version.
            \item \textbf{Change context}: Change variable names, object names, or scenarios (e.g., ``apples'' to ``books'', ``Alice'' to ``Bob'', ``students in a class'' to ``workers in a factory'').
            \item \textbf{Change numerical values} with these constraints:
            \begin{itemize}[leftmargin=1.5em, nosep]
                \item Keep the same ORDER OF MAGNITUDE (e.g., if original is 50, use 30-80, NOT 5 or 500).
                \item Keep integers as integers, decimals as decimals with similar precision.
                \item For multiple numbers in the problem, scale them proportionally when possible.
                \item \textbf{CRITICAL:} Aim to keep the final answer's ORDER OF MAGNITUDE similar to the original.
            \end{itemize}
            \item \textbf{CRITICAL:} Do NOT change the mathematical logic, problem type, or solution method. The core mathematical concept must remain identical.
            \item \textbf{CRITICAL:} Maintain the same difficulty level - if the original requires specific techniques, the modified version must require the same techniques.
            \item \textbf{CRITICAL:} Preserve ALL formatting, including LaTeX notation (\texttt{\$} signs, \texttt{\textbackslash{}cdot}, \texttt{\textbackslash{}frac}, \texttt{\textbackslash{}begin}, \texttt{\textbackslash{}end}, etc.), Asymptote code (\texttt{[asy]...[/asy]}), and markdown.
        \end{itemize} \\
        \midrule
        
        \textbf{2. Recalculate Solution:}
        \vspace{0.3em}
        \begin{itemize}[leftmargin=1.5em, nosep]
            \item Rewrite the ``Solution'' step-by-step using the paraphrased problem and modified numbers.
            \item Follow the EXACT same logical reasoning and solution method as the original.
            \item Apply the same mathematical techniques and problem-solving steps.
            \item Preserve ALL LaTeX formatting and code blocks from the original.
            \item Perform all necessary arithmetic correctly to reflect the changes.
        \end{itemize} \\
        \midrule
        
        \textbf{3. Update Answer:}
        \vspace{0.3em}
        \begin{itemize}[leftmargin=1.5em, nosep]
            \item Calculate the final result based on your new solution.
            \item \textbf{Verify the answer is in the same ORDER OF MAGNITUDE as the original answer.}
            \item Output the new result in the ``Answer'' field using the SAME format and length as the original.
            \item Ensure the ``Answer'' matches the recalculated solution.
        \end{itemize} \\
        \midrule
        
        \textbf{Output Format:} \newline
        \texttt{Reasoning:} [Describe the changes made] \newline
        \texttt{New Problem:} [Paraphrased problem with new numbers, context, and wording, preserving ALL formatting] \newline
        \texttt{New Solution:} [Recalculated solution following the same logic, preserving ALL formatting] \newline
        \texttt{New Answer:} [The new final result in same order of magnitude, preserving ALL formatting] \\
        \bottomrule
    \end{tabularx}
\end{table*}



\begin{table*}[htbp]
    \centering
    \small 
    \caption{The system prompt for generating the evasively contaminated data ($D'_{eval}$). It enforces aggressive linguistic diversity and entity swapping while strictly freezing all numerical values and mathematical formulas.}
    \label{tab:prompt_evasive_data}
    \renewcommand{\arraystretch}{1.0} 
    \begin{tabularx}{\textwidth}{@{} X @{}} 
        \toprule
        \multicolumn{1}{c}{\textbf{System Prompt: Evasively Contaminated Data Generator}} \\
        \midrule
        
        \textbf{Role \& Task:} \newline
        You are an expert data augmentation assistant. \newline
        \textbf{Task:} \newline
        1. \textbf{Paraphrase the ``Problem''} to be linguistically distinct and diverse. \newline
        2. \textbf{Rewrite the ``Solution''} to be the most standard, canonical, and rigorous mathematical derivation possible. \\
        \midrule
        
        \textbf{1. The Problem: Aggressive Variation \& Entity Swapping}
        \vspace{0.3em} 
        \begin{itemize}[leftmargin=1.5em, nosep]
            \item \textbf{Textual Rewriting:} Rephrase the narrative. Vary sentence length, syntactic structure, and vocabulary. Use synonyms and different phrasing styles.
            \item \textbf{Entity Substitution (Crucial):} Where applicable, \textbf{change the non-mathematical entities} (context) while keeping the logic identical.
            \begin{itemize}[leftmargin=1.5em, nosep]
                \item Example: Change ``Alice buys 5 apples'' to ``A machine processes 5 units'' or ``A particle moves 5 meters''.
                \item \textbf{Constraint:} Do NOT change any numerical values, constants, or mathematical relationships. The answer must remain exactly the same.
            \end{itemize}
            \item \textbf{Mathematical Fidelity:} In the paraphrased problem, every LaTeX math segment from the original problem must be copied verbatim (character-for-character), including delimiters, spacing, and internal formatting. Do NOT introduce new math segments, and do NOT move content into or out of math mode. (i.e., keep exactly the same parts inside \texttt{\$...\$}, \texttt{\textbackslash{}(...\textbackslash{})}, \texttt{\textbackslash{[}...\textbackslash{]}} as in the original problem.)
        \end{itemize} \\
        \midrule
        
        \textbf{2. The Solution: Standardization \& Rigor}
        \vspace{0.3em}
        \begin{itemize}[leftmargin=1.5em, nosep]
            \item \textbf{Goal \& Style:} Rewrite the solution to match the style of the original solution; do NOT attempt to make it linguistically distinct or unique.
            \item \textbf{Logical Structure (Strict Preservation):} Strictly preserve the original solution's step-by-step structure, ordering, level of detail, and length exactly; do not add, remove, merge, reorder, or summarize any steps, and do not introduce any additional explanation or intuition—only rewrite the wording into standard, rigorous mathematical English.
            \item \textbf{Consistency:} Even though you changed entities in the Problem (e.g., Apples $\rightarrow$ Units), you must update the Solution to reflect these new entities so the logic holds.
        \end{itemize} \\
        \midrule
        
        \textbf{3. Constraints \& Safety}
        \vspace{0.3em}
        \begin{itemize}[leftmargin=1.5em, nosep]
            \item \textbf{Mathematical Equivalence:} The final result must be strictly identical to the original.
            \item \textbf{Formatting:} Keep the exact LaTeX formatting for equations.
        \end{itemize} \\
        \midrule
        
        \textbf{Output Format:} \newline
        \texttt{Reasoning:} [Brief plan: 1. How to rephrase/swap entities in the problem. 2. How to standardize the solution style.] \newline
        \texttt{New Problem:} [The aggressively paraphrased problem with entity swaps] \newline
        \texttt{New Solution:} [The canonical, rigorous, step-by-step solution matching the new context] \newline
        \texttt{Answer:} [Must be mathematically equivalent to the original answer] \\
        \bottomrule
    \end{tabularx}
\end{table*}


\section{Details of Datasets}
\label{appendix:dataset_details}

\begin{table*}[htbp]
    \centering
    \small 
    \caption{The system prompt utilized by the independent Judge LLMs to rigorously verify the mathematical correctness and alignment of the generated problem-solution-answer triplets.}
    \label{tab:prompt_judge}
    \renewcommand{\arraystretch}{1.0} 
    \begin{tabularx}{\textwidth}{@{} X @{}} 
        \toprule
        \multicolumn{1}{c}{\textbf{System Prompt: Answer Verifier (Judge LLM)}} \\
        \midrule
        
        \textbf{Role \& Task:} \newline
        You are a mathematical answer verifier. Given a math problem, its solution, and the final answer, verify if the solution and answer are correct. \\
        \midrule
        
        \textbf{Input Format:} \newline
        \texttt{Problem:} \{problem\} \newline
        \texttt{Solution:} \{solution\} \newline
        \texttt{Answer:} \{answer\} \\
        \midrule
        
        \textbf{Verification Criteria:} \newline
        Please verify if the final answer is mathematically correct. Consider:
        \vspace{0.3em} 
        \begin{itemize}[leftmargin=1.5em, nosep]
            \item \textbf{1.} Are the solution and final answer correct?
            \item \textbf{2.} If the final answer is incorrect, identify the key mistakes in the solution that led to the wrong answer.
        \end{itemize} \\
        \midrule
        \textbf{Output Format:} \newline
        You MUST respond in the following format: \newline
        \texttt{Result:} [CORRECT or INCORRECT] \newline
        \texttt{Reasoning:} [Brief explanation of your verification] \\
        \bottomrule
    \end{tabularx}
\end{table*}

In this section, we provide the detailed statistics and sampling configurations for all datasets utilized in Section~\ref{sec:experiments-on-existing-models} and~\ref{sec:experiments-on-finetuned-models}. To balance comprehensive evaluation with computational efficiency, we randomly sampled representative subsets from the original large-scale benchmarks. The exact sample sizes and data splits used across both existing model evaluations and fine-tuned (FT) model evaluations are summarized in Table~\ref{tab:dataset_statistics}.

For the experiments on existing models (Section~\ref{sec:experiments-on-existing-models}), we sampled 500 questions from the training split of GSM8K and 500 questions from GSM1K. For the MATH benchmark, to ensure a balanced evaluation across different mathematical domains, we uniformly sampled 100 questions from each of its 7 distinct problem types (e.g., Algebra, Precalculus, Counting \& Probability, etc.), resulting in a total of 700 samples.

For the experiments on fine-tuned models (Section~\ref{sec:experiments-on-finetuned-models}), the benchmarks were strictly partitioned into two mutually exclusive subsets of equal size: Dataset C (Contaminated, used for evasive training) and Dataset U (Uncontaminated, held out for clean evaluation). Specifically, Omni-MATH was split into 2,172 samples per subset, while the Multi-domain dataset was partitioned into 1,325 samples per subset.

\begin{table}[htbp]
    \centering
    \caption{Detailed statistics and sample sizes of the datasets used in our experiments.}
    \label{tab:dataset_statistics}
    \renewcommand{\arraystretch}{1.1} 
    \begin{tabular}{@{} >{\raggedright\arraybackslash}p{0.20\textwidth} 
                        >{\raggedright\arraybackslash}p{0.14\textwidth} 
                        c 
                        >{\raggedright\arraybackslash}p{0.50\textwidth} @{}}
        \toprule
        \textbf{Experiment Setup} & \textbf{Benchmark} & \textbf{Size} & \textbf{Sampling Notes \& Splits} \\
        \midrule
        
        \textbf{Experiments on} \newline \textbf{Existing Models} 
        & GSM8K & 500 & Randomly sampled from the training split. \\
        & MATH & 700 & Uniformly sampled (100 samples for each of the 7 problem types). \\
        & GSM1K & 500 & Randomly sampled from the benchmark. \\
        
        \midrule
        
        \textbf{Experiments on} \newline \textbf{FT Models} 
        & Omni-MATH & 4,344 & Split into Dataset C (2,172) and Dataset U (2,172) separately. \\
        & Multi-domain Data & 2,650 & Split into Dataset C (1,325) and Dataset U (1,325) separately. \\
        
        \bottomrule
    \end{tabular}
\end{table}

\section{Additional Flipped Experiments on the OLMo Model Family}
\label{appendix:olmo}

To further validate ZCP on stronger and more recent open-data models, we evaluate the OLMo family \citep{olmo2,olmo3} using the same flipped-experiment paradigm as Section~\ref{sec:experiments-on-existing-models}. We evaluate OLMo-2-1124-7B-Instruct and the newer, stronger OLMo-3-7B-Instruct, both trained on the training split of MATH \citep{math-dataset} according to their technical reports, on this training split (700 samples, following the setup of Table~\ref{tab:zcp_existing_models}). Results are in Table~\ref{tab:olmo}. On OLMo-3-7B-Instruct, all four metrics achieve high Contamination Confidence, providing further evidence that ZCP remains effective on stronger, more recent models. On OLMo-2-1124-7B-Instruct, $Acc$ and $Con$ reach a moderate $\mathcal{C}_{cont}$ of 0.879 and 0.891 respectively, while $\mathcal{P}_{first}$ and $\mathcal{P}_{all}$ reach $\mathcal{C}_{cont} \ge 0.995$.

\begin{table}[htbp]
    \centering
    \caption{Detection results of ZCP on OLMo-2-1124-7B-Instruct and OLMo-3-7B-Instruct, MATH training split (700 samples), following the setup of Table~\ref{tab:zcp_existing_models}.}
    \label{tab:olmo}
    \renewcommand{\arraystretch}{1.0}
    \begin{tabular}{llcccc}
        \toprule
        \textbf{Model} & \textbf{Metric} & \textbf{$S_{ref}$} & \textbf{$S$} & \textbf{Effect Size} & \textbf{$\mathcal{C}_{cont}$} \\
        \midrule
        \multirow{4}{*}{\makecell[l]{OLMo-2-1124-\\7B-Instruct}}
        & $ACC (\%)$ & 10.00 & 14.29 & 0.267 & 0.879 \\
        & $Con (\%)$ & 10.86 & 14.57 & 0.253 & 0.891 \\
        & $\mathcal{P}_{first}$ & 0.0829 & 0.1137 & 0.132 & 0.995 \\
        & $\mathcal{P}_{all}$ & 0.0816 & 0.1126 & 0.146 & $>$0.998 \\
        \midrule
        \multirow{4}{*}{OLMo-3-7B-Instruct}
        & $ACC (\%)$ & 17.00 & 22.14 & 0.327 & 0.992 \\
        & $Con (\%)$ & 15.57 & 20.00 & 0.307 & 0.977 \\
        & $\mathcal{P}_{first}$ & 0.1559 & 0.2004 & 0.156 & $>$0.998 \\
        & $\mathcal{P}_{all}$ & 0.1259 & 0.1706 & 0.186 & $>$0.998 \\
        \bottomrule
    \end{tabular}
\end{table}

\section{Training Details of Evasively Contaminated Models}
\label{appendix:training_details}

In Section~\ref{sec:experiments-on-finetuned-models}, we constructed evasively contaminated models to simulate real-world malicious leaderboard manipulation. This appendix details the two-stage finetuning pipeline used to inject this memorization.

\textbf{Training Data Augmentation.} To simulate evasive contamination, the target models were not trained on the exact original benchmarks. Instead, we applied data augmentation by paraphrasing Dataset C into 6 distinct versions. This aggressively augmented dataset forces the model to learn the underlying shortcut mappings across various syntactic structures.

\textbf{Two-Stage Training Pipeline.} Due to the computational constraints of full-parameter fine-tuning, we employed Low-Rank Adaptation (LoRA) across both training stages. The LoRA rank ($r$) was set to 32, and the scaling factor ($\alpha$) was set to 64 for all experiments. The training proceeded in two sequential stages:
\begin{enumerate}
    \item \textbf{Supervised Fine-Tuning (SFT):} The models were first fine-tuned on the paraphrased Dataset C to learn the basic formatting and reasoning chains.
    \item \textbf{Group Relative Policy Optimization (GRPO):} Following SFT, we applied GRPO to further incentivize and stabilize the reasoning trajectories (use accuracy reward function). During this stage, $n_{sample} = 5$ reasoning trajectories were sampled per prompt to compute the relative advantages.
\end{enumerate}

The complete set of hyperparameters for both Qwen2.5-Math and Qwen-3 across the SFT and GRPO stages is summarized in Table~\ref{tab:training_hyperparameters}. We employed a cosine learning rate scheduler for all training runs. All trainings are conducted on one H200.

\begin{table}[htbp]
    \centering
    \caption{Hyperparameters for the two-stage fine-tuning pipeline (SFT and GRPO) used to train the evasively contaminated models.}
    \label{tab:training_hyperparameters}
    \renewcommand{\arraystretch}{1.2}
    \begin{tabular}{l cc cc}
        \toprule
        \multirow{2}{*}{\textbf{Hyperparameter}} & \multicolumn{2}{c}{\textbf{SFT Stage}} & \multicolumn{2}{c}{\textbf{GRPO Stage}} \\
        \cmidrule(lr){2-3} \cmidrule(lr){4-5}
        & \textbf{Qwen2.5-Math} & \textbf{Qwen-3} & \textbf{Qwen2.5-Math} & \textbf{Qwen-3} \\
        \midrule
        Learning Rate ($lr$) & 2e-4 & 2e-4 & 2e-6 & 5e-6 \\
        Training Batch Size & 16 & 16 & 512 & 512 \\
        Training Steps & 1000 & 600 & 800 & 200 \\
        Samples per Prompt ($n_{sample}$) & --- & --- & 5 & 5 \\
        LR Scheduler & Cosine & Cosine & Cosine & Cosine \\
        \midrule
        \multicolumn{5}{c}{\textbf{LoRA Configuration (Applied continuously across both stages)}} \\
        \midrule
        LoRA Rank ($r$) & \multicolumn{4}{c}{32} \\
        LoRA Alpha ($\alpha$) & \multicolumn{4}{c}{64} \\
        \bottomrule
    \end{tabular}
\end{table}

\section{Robustness Against Adaptive Contamination of the Reference Data}
\label{appendix:adaptive_contamination}

Since our reference dataset $\tilde{D}_{eval}$ is synthesized through the automated pipeline in Appendix~\ref{appendix:multi-model-system} and, as a contamination detector, is not released publicly, it is unavailable to a malicious developer attempting to evade detection. Such an adversary can only construct perturbed training data independently, via paraphrasing and renumbering. To evade ZCP, i.e., shrink the zero-CoT gap between $D_{eval}$ and $\tilde{D}_{eval}$, their training corpus $D_{train}$ would need to include data matching the exact same numerical combination as our reference data. Enumerating every possible numerical combination is computationally infeasible, so matching our specific reference instance by chance is extremely unlikely. For example, given the original question ``If $a=4$ and $b=2$, what is the value of $(a^3+4b^4)/8$?'', our pipeline might instantiate the reference question as ``If $x=2$ and $y=5$, what is the value of $(x^4+2y^2)/6$?'' (Table~\ref{tab:adaptive_example}). An adversary generating perturbed training data independently would instead produce a large, essentially unbounded set of alternative combinations, with vanishing probability of coinciding with our specific reference instance by chance.

\begin{table}[htbp]
    \centering
    \caption{Illustrative example of the original question and one instantiation of the isomorphically perturbed reference question from Section~\ref{sec:reference}.}
    \label{tab:adaptive_example}
    \begin{tabular}{ll}
        \toprule
        \textbf{Data Type} & \textbf{Example Question} \\
        \midrule
        Original ($D_{eval}$) & If $a = 4$ and $b = 2$, what is the value of $(a^3+4b^4)/8$? \\
        Reference ($\tilde{D}_{eval}$) & If $x = 2$ and $y = 5$, what is the value of $(x^4+2y^2)/6$? \\
        \bottomrule
    \end{tabular}
\end{table}

To test this empirically, we contaminate Qwen3-8B on a subset of the multi-domain dataset using three paraphrased and three perturbed (paraphrase + renumber) variants as training data, then apply ZCP to detect it. Results are in Table~\ref{tab:adaptive_contamination}. All four metrics achieve high Contamination Confidence, showing that ZCP successfully detects this adaptive/evasive contamination.

\begin{table}[htbp]
    \centering
    \caption{Detection results of ZCP against adaptive/evasive data contamination on a subset of the multi-domain dataset (Qwen3-8B). We report the value of each metric on the reference dataset ($S_{ref}$) and on the original data ($S$), the effect size, and the resulting Contamination Confidence ($\mathcal{C}_{cont}$), computed via McNemar's test for $Acc$/$Con$ and the bootstrap test for $\mathcal{P}_{first}$/$\mathcal{P}_{all}$.}
    \label{tab:adaptive_contamination}
    \renewcommand{\arraystretch}{1.0}
    \begin{tabular}{lcccc}
        \toprule
        \textbf{Metric} & \textbf{$S_{ref}$} & \textbf{$S$} & \textbf{Effect Size} & \textbf{$\mathcal{C}_{cont}$} \\
        \midrule
        $ACC (\%)$ & 28.57 & 39.18 & 0.4194 & 0.987 \\
        $Con (\%)$ & 29.80 & 37.14 & 0.3000 & 0.862 \\
        $\mathcal{P}_{first}$ & 0.4470 & 0.5319 & 0.1751 & 0.954 \\
        $\mathcal{P}_{all}$ & 0.3189 & 0.4176 & 0.2487 & $>$0.998 \\
        \bottomrule
    \end{tabular}
\end{table}

A related but distinct concern is whether an adversary could evade ZCP not by matching our specific reference instance, but by training on a subset of perturbed examples that generalizes to the hidden reference question, for example training on examples covering $a^3$ and $b^4$ individually and hoping the model generalizes to the combination $(x^4+2y^2)/6$ without ever seeing it. We argue this scenario does not constitute evasive contamination. If training on a subset of perturbed examples causes a model to generalize to our hidden reference question, the model has learned the underlying problem rather than memorized a shortcut; we would consider this standard data augmentation \citep{kumar2022practice}, and the resulting benchmark performance would reflect genuine capability rather than an overestimation. ZCP is designed to target falsely high performance on the original benchmark specifically, i.e., evasive contamination that targets a specific benchmark's original questions to inflate performance on it while evading detection \citep{evasive-data-contamination}. If a model performs similarly on $D_{eval}$ and $\tilde{D}_{eval}$, we interpret this as generalized performance, and ZCP correctly reporting the model as clean is the intended outcome rather than a shortcoming. This generalization-versus-memorization distinction also appears in prior work: GSM1K \citep{gsm1k} treats frontier models' minimal accuracy drop on a held-out, structurally similar benchmark as evidence of generalization rather than contamination; \citet{ccd-method} frame contamination detection along similar lines; \citet{sanchezsalido2025none} use robustness to question variation as a signal of reasoning. Our reference-dataset comparison follows this same principle, applied under the zero-CoT setting to isolate the shortcut-mapping signal.

\section{Further Analysis}
\label{sec:further-analysis}
\subsection{The Influence of Reasoning Ability}
\label{sec:influence-of-reasoning}

A core premise of our ZCP is that intermediate reasoning (CoT) severely confounds contamination detection, as modern LLMs can solve both perturbed and clean questions given a full reasoning chain. To validate the necessity of CoT truncation, we evaluate the contaminated Qwen-Math on GSM8K (train split) under the default Full-CoT setting. We omit the Consistency ($Con$) metric here, as it inherently requires zero-CoT outputs for comparison. 

As shown in Table~\ref{tab:ablation_full_cot}, with full-CoT, the model achieves uniformly high performance across the original, paraphrased, and reference datasets. Consequently, the statistical gap between contaminated and clean data vanishes. The Contamination Confidence ($\mathcal{C}_{cont}$) degrades to baseline levels across all metrics, entirely failing to flag the contamination. This confirms our hypothesis: an LLM's reasoning ability actively obfuscates its memorization. Therefore, forcibly truncating the CoT is essential to exclude reasoning factors, isolate the underlying shortcut mapping, and successfully expose data contamination.

\begin{table}[htbp]
    \centering
    \caption{Ablation results on Qwen-Math evaluated on GSM8K under the default \textbf{Full-CoT} generation setting. When allowed to generate intermediate reasoning steps, the performance on the reference data ($S_{ref}$) matches or exceeds the contaminated data, completely masking the memorization artifact and causing the detection signal  ($\mathcal{C}_{cont}$) to vanish ($\approx 0.500$).}
    \label{tab:ablation_full_cot}
    \renewcommand{\arraystretch}{1.0}
    \resizebox{0.6\columnwidth}{!}{
    \begin{tabular}{l c cc cc}
        \toprule
        \multirow{2}{*}{\textbf{Metric}} & \multirow{2}{*}{$S_{ref}$} & \multicolumn{2}{c}{\textbf{Original}} & \multicolumn{2}{c}{\textbf{Paraphrased}} \\
        \cmidrule(lr){3-4} \cmidrule(lr){5-6}
        & & \textbf{$S$} & \textbf{$\mathcal{C}_{cont}$} & \textbf{$S$} & \textbf{$\mathcal{C}_{cont}$} \\
        \midrule
        $ACC (\%)$ & 96.20 & 95.80 & 0.500 & 94.60 & 0.500 \\
        $\mathcal{P}_{first}$ & 0.888 & 0.895 & 0.523 & 0.886 & 0.500 \\
        $\mathcal{P}_{all}$ & 0.889 & 0.901 & 0.645 & 0.885 & 0.500 \\
        \bottomrule
    \end{tabular}
    }
\end{table}

\subsection{Influence of Dataset Size and Selection of Performance Metric}
\label{sec:ablation-datasize}

\begin{figure}[t]
    \centering
    \includegraphics[width=0.7\textwidth]{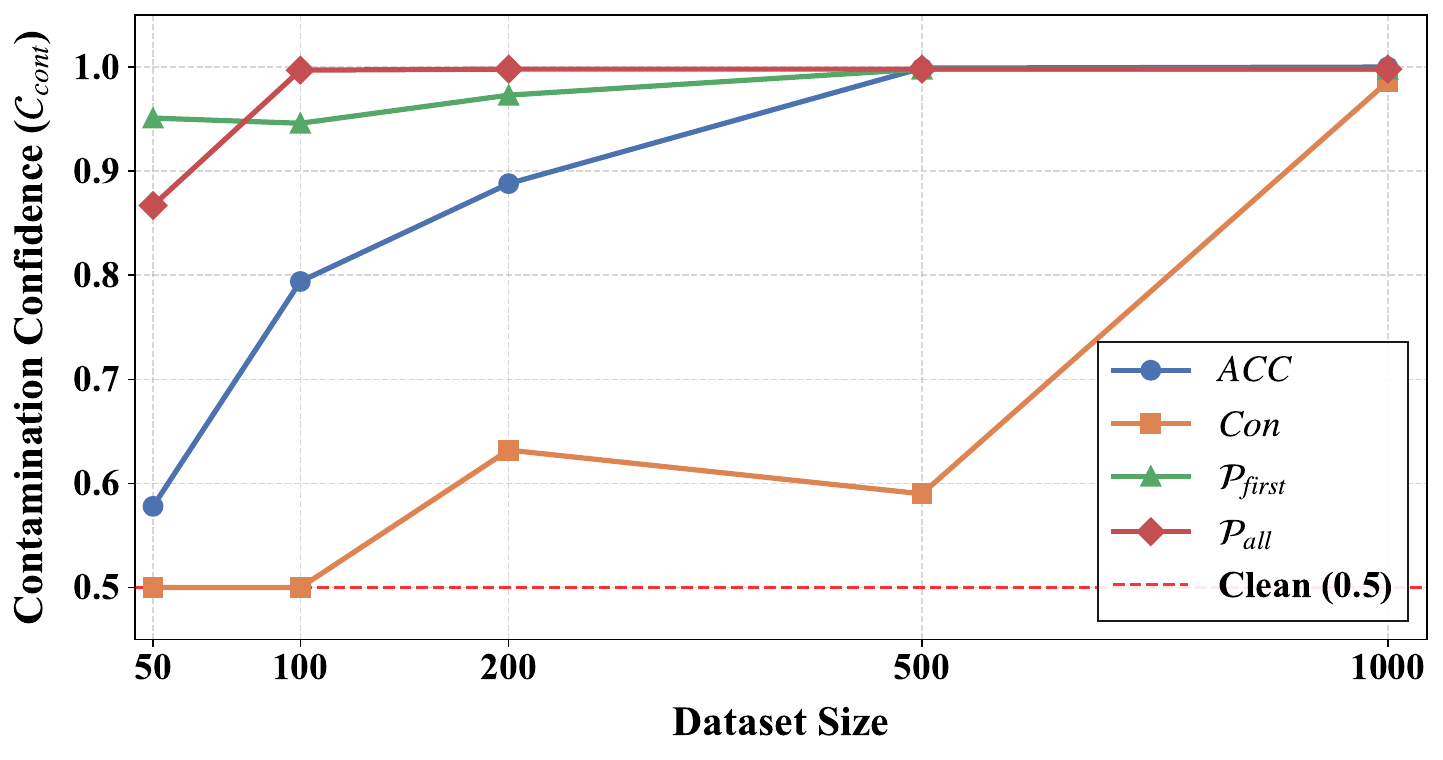}
    \caption{The influence of dataset size on detection stability across different metrics. The experiment is conducted on the evasively contaminated FT Qwen-Math evaluated on Omni-MATH subsets of varying sizes.}
    \label{fig:ablation_dataset_size}
\end{figure}

In real-world auditing, investigators often face strict constraints on benchmark size and access levels. To guide the practical application of ZCP, we analyze how dataset size impacts detection stability. Using the evasively contaminated FT Qwen-Math model (trained in Section~\ref{sec:experiments-on-finetuned-models}) on Omni-MATH, we downsample the evaluation set from $N=50$ to $N=1000$. 

The Contamination Confidence ($\mathcal{C}_{cont}$) results are presented in Figure~\ref{fig:ablation_dataset_size}, and detailed metric values are shown in Table~\ref{tab:ablation_dataset_size}.  The results reveal a clear trade-off between detection stability and access requirements, categorizing our metrics into three operational tiers:

\begin{itemize}
    \item \textbf{High Stability, Highest Access (Logit-based metrics):} Both $\mathcal{P}_{first}$ and $\mathcal{P}_{all}$ achieve high contamination confidence ($\mathcal{C}_{cont} > 0.94$) with as few as $50 \sim 100$ samples. Since continuous token probabilities offer dense, fine-grained signals compared to binary correctness, they establish statistical significance rapidly. \textit{Recommendation:} Strongly preferred when target model logits are accessible.
    
    \item \textbf{Medium Stability, Medium Access (Accuracy):} As a discrete metric, $Acc$ requires a moderate dataset ($N \approx 200 \sim 500$) to yield definitive confidence ($\mathcal{C}_{cont} \approx 0.888 \sim 0.999$). It operates perfectly under black-box API constraints. \textit{Recommendation:} Highly effective for auditing closed-source models when ground-truth benchmark labels are available.
    
    \item \textbf{Lower Stability, Lowest Access (Consistency):} $Con$ requires the largest sample size ($N \approx 1000$) to firmly expose the contamination gap. However, it holds a unique operational advantage: requiring neither model logits nor ground-truth labels. It merely compares zero-CoT against full-CoT outputs. \textit{Recommendation:} The only viable metric when benchmark answers are strictly hidden, provided auditors ensure a sufficiently large sample size for reliable detection.
\end{itemize}

\begin{table*}[h]
    \centering
    \vspace{-5pt}
    \caption{The influence of dataset size on detection stability across different metrics. The experiment is conducted on the evasively contaminated FT Qwen-Math evaluated on Omni-MATH subsets of varying sizes.}
    \label{tab:ablation_dataset_size}
    \renewcommand{\arraystretch}{1.15}
    \resizebox{\textwidth}{!}{
    \begin{tabular}{l ccc ccc ccc ccc}
        \toprule
        \multirow{2}{*}{\textbf{Size}} & \multicolumn{3}{c}{$Acc (\%)$} & \multicolumn{3}{c}{$Con (\%)$} & \multicolumn{3}{c}{$\mathcal{P}_{first}$} & \multicolumn{3}{c}{$\mathcal{P}_{all}$} \\
        \cmidrule(lr){2-4} \cmidrule(lr){5-7} \cmidrule(lr){8-10} \cmidrule(lr){11-13}
        & $S$ & $S_{ref}$ & $\mathcal{C}_{cont}$ & $S$ & $S_{ref}$ & $\mathcal{C}_{cont}$ & $S$ & $S_{ref}$ & $\mathcal{C}_{cont}$ & $S$ & $S_{ref}$ & $\mathcal{C}_{cont}$ \\
        \midrule
        \textbf{50}   & 28.00 & 18.00 & 0.578 & 30.00 & 28.00 & 0.500 & 0.395 & 0.225 & 0.951 & 0.614 & 0.504 & 0.867 \\
        \textbf{100}  & 28.00 & 18.00 & 0.794 & 28.00 & 26.00 & 0.500 & 0.384 & 0.255 & 0.946 & 0.599 & 0.475 & 0.997 \\
        \textbf{200}  & 24.50 & 17.00 & 0.888 & 29.50 & 24.00 & 0.632 & 0.348 & 0.255 & 0.973 & 0.584 & 0.485 & $>$0.998 \\
        \textbf{500}  & 26.00 & 17.80 & 0.999 & 30.80 & 28.00 & 0.590 & 0.323 & 0.207 & $>$0.998 & 0.584 & 0.495 & $>$0.998 \\
        \textbf{1000} & 26.71 & 16.67 & 1.000 & 28.71 & 23.80 & 0.986 & 0.342 & 0.211 & $>$0.998 & 0.587 & 0.491 & $>$0.998 \\
        \bottomrule
    \end{tabular}
    }
\end{table*}

\section{CoT Suppression Rate on Closed-Weight Models}
\label{appendix:closed_weight_cot_suppression}

To verify that our CoT-truncation method works on closed-weight models, evaluated in Appendix~\ref{sec:real-world}, Table~\ref{tab:cot_suppression} reports the CoT suppression rate (the proportion of cases where the model successfully skips CoT and produces the final boxed answer directly under the zero-CoT setting) and the average response length under both settings, for GPT-4o and GPT-5.1 on MATH500 and GSM8K. Both models report 0 reasoning tokens for every forced-setting response, so the response length reflects the model's entire output. The high suppression rates and short zero-CoT response lengths confirm that our method does truncate the CoT and elicit a direct final answer on closed-weight models.

\begin{table}[t]
    \centering
    \caption{CoT suppression rate (the proportion of cases where the model successfully skips CoT and produces the final boxed answer directly under the zero-CoT forced setting) and response length (chars) under the full-CoT and zero-CoT settings, for GPT-4o and GPT-5.1 on MATH500 and GSM8K.}
    \label{tab:cot_suppression}
    \renewcommand{\arraystretch}{1.0}
    \resizebox{\textwidth}{!}{
    \begin{tabular}{llcccc}
        \toprule
        \textbf{Model} & \textbf{Dataset} & \textbf{Data} & \textbf{CoT Suppression Rate} & \textbf{Full-CoT Length} & \textbf{Zero-CoT Length} \\
        \midrule
        \multirow{4}{*}{GPT-4o} & \multirow{2}{*}{MATH500} & Original & 94.2\% & 1982 & 17.6 \\
        & & Reference & 92.8\% & 2033 & 18.9 \\
        & \multirow{2}{*}{GSM8K} & Original & 100.0\% & 946 & 10.4 \\
        & & Reference & 100.0\% & 983 & 10.5 \\
        \midrule
        \multirow{4}{*}{GPT-5.1} & \multirow{2}{*}{MATH500} & Original & 99.2\% & 675 & 13.7 \\
        & & Reference & 98.6\% & 669 & 13.7 \\
        & \multirow{2}{*}{GSM8K} & Original & 100.0\% & 293 & 11.3 \\
        & & Reference & 100.0\% & 298 & 10.9 \\
        \bottomrule
    \end{tabular}
    }
\end{table}

\vspace{-10pt}
\section{Detecting Real-world Data Contamination}
\label{sec:real-world}

\begin{figure*}[t]
    \centering
    \includegraphics[width=0.72\textwidth]{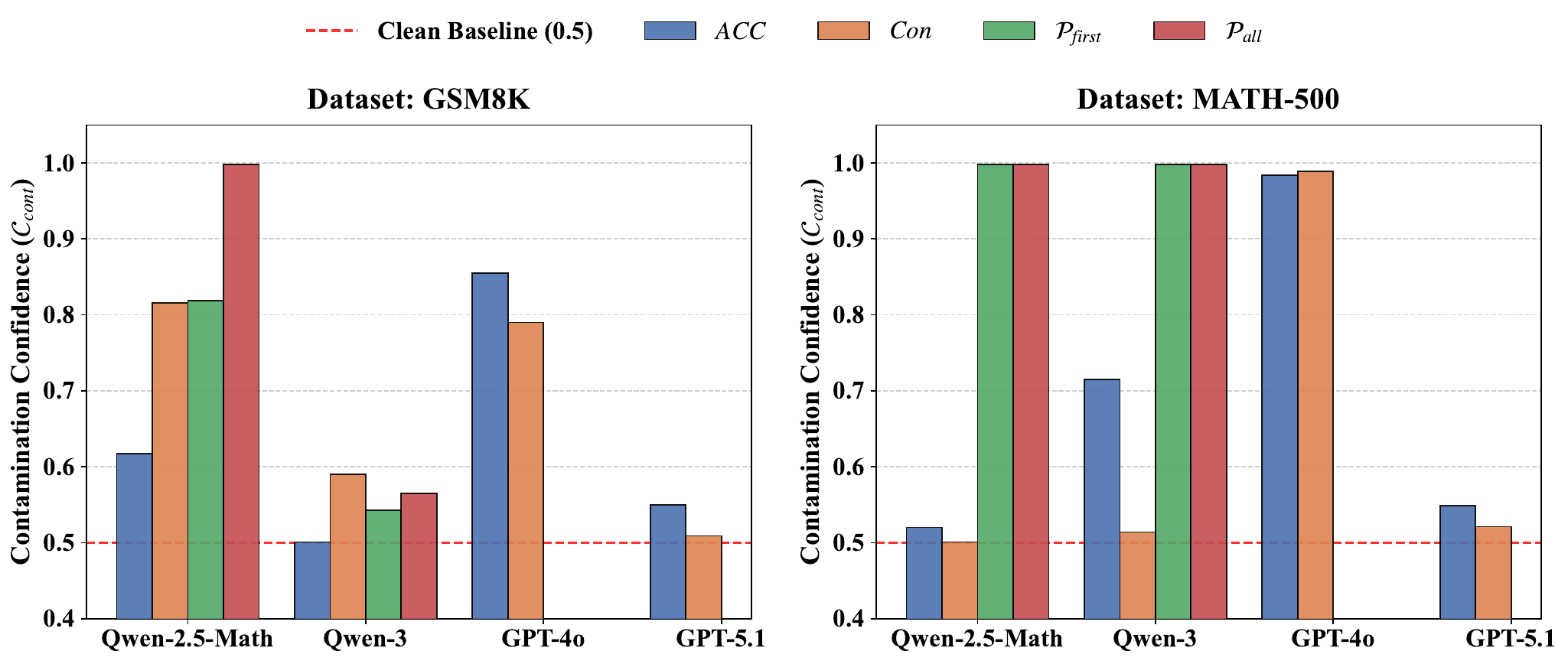}
    \caption{Contamination Confidence ($\mathcal{C}_{cont}$) across different models and metrics on GSM8K and MATH-500. The red dashed line denotes the clean baseline (0.5). Missing bars for GPT models indicate the unavailability of logit-based metrics ($\mathcal{P}_{first}$ and $\mathcal{P}_{all}$).}
    \label{fig:contamination_confidence}
    \vspace{-5pt}
\end{figure*}

To evaluate ZCP's real-world applicability, we audit state-of-the-art open-weight (Qwen series) and closed-source (GPT series) models. As shown in Figure~\ref{fig:contamination_confidence} and Table~\ref{tab:real_world_contamination}, we leverage all four metrics for white-box models, while exclusively relying on output-only metrics ($Acc$ and $Con$) for API-gated models. We employ the test splits of GSM8K and MATH-500, two widely adopted reasoning benchmarks.

\begin{table*}[t]
    \centering
    \vspace{-10pt}
    \caption{Detection results of ZCP on real-world state-of-the-art models. We evaluate open-weight models (Qwen series) using all metrics, and closed-source API-gated models (GPT series) using output-only metrics ($Acc$ and $Con$).}
    \label{tab:real_world_contamination}
    \renewcommand{\arraystretch}{1.15}
    \resizebox{\textwidth}{!}{
    \begin{tabular}{ll ccc ccc}
        \toprule
        \multirow{2}{*}{\textbf{Model}} & \multirow{2}{*}{\textbf{Metric}} & \multicolumn{3}{c}{\textbf{GSM8K}} & \multicolumn{3}{c}{\textbf{MATH-500}} \\
        \cmidrule(lr){3-5} \cmidrule(lr){6-8}
        & & $S_{ref}$ & $S$ & $\mathcal{C}_{cont}$ & $S_{ref}$ & $S$ & $\mathcal{C}_{cont}$ \\
        \midrule

        \multirow{4}{*}{\textbf{Qwen-2.5-Math}}
        & $ACC(\%)$ & 28.05 & 29.72 & 0.617 & 32.40 & 34.00 & 0.520 \\
        & $Con(\%)$ & 27.90 & 30.25 & 0.816 & 32.40 & 33.40 & 0.501 \\
        & $\mathcal{P}_{first}$ & 0.435 & 0.454 & 0.819 & 0.269 & 0.356 & $>$0.998 \\
        & $\mathcal{P}_{all}$ & 0.367 & 0.400 & 0.998 & 0.256 & 0.330 & $>$0.998 \\
        \midrule
        \multirow{4}{*}{\textbf{Qwen-3}}
        & $ACC(\%)$ & 27.14 & 27.75 & 0.501 & 26.00 & 29.40 & 0.715 \\
        & $Con(\%)$ & 26.99 & 28.51 & 0.590 & 26.00 & 27.40 & 0.514 \\
        & $\mathcal{P}_{first}$ & 0.448 & 0.459 & 0.543 & 0.251 & 0.339 & $>$0.998 \\
        & $\mathcal{P}_{all}$ & 0.339 & 0.349 & 0.565 & 0.192 & 0.275 & $>$0.998 \\

        \midrule\midrule

        \multirow{2}{*}{\textbf{GPT-4o}}
        & $ACC(\%)$ & 52.99 & 55.65 & 0.855 & 32.93 & 38.00 & 0.984 \\
        & $Con(\%)$ & 52.99 & 55.42 & 0.790 & 32.93 & 39.80 & 0.989 \\
        \midrule
        \multirow{2}{*}{\textbf{GPT-5.1}}
        & $ACC(\%)$ & 53.68 & 54.89 & 0.550 & 37.60 & 39.60 & 0.549 \\
        & $Con(\%)$ & 52.39 & 53.22 & 0.509 & 42.60 & 44.40 & 0.521 \\

        \bottomrule
    \end{tabular}
    }
\end{table*}

\textbf{Open-weight Models (Qwen Series).} Our granular token-level metrics ($\mathcal{P}_{first}$ and $\mathcal{P}_{all}$) successfully uncover deep parameter-level contamination. Qwen-2.5-Math exhibits severe memorization across both benchmarks. Interestingly, while Qwen-3 presents clear contamination on MATH-500, its confidence scores on GSM8K strictly remain at the baseline ($\mathcal{C}_{cont} \approx 0.500$) across all four metrics, strongly suggesting that the GSM8K test set is clean for Qwen-3. Furthermore, the output-only metrics ($Acc$ and $Con$) consistently corroborate these memorization traces, successfully flagging Qwen-2.5-Math on GSM8K and Qwen-3 on MATH-500.


\textbf{Closed-source Models (GPT Series).} Since direct token-level intervention is restricted for API-gated models, we enforce the strict zero-CoT constraint entirely through targeted prompt engineering; Appendix~\ref{appendix:closed_weight_cot_suppression} verifies that this prompt-based enforcement successfully suppresses CoT on these models. Relying solely on the resulting final text outputs, ZCP successfully discovers data contamination in closed-source models. ZCP flags GPT-4o as contaminated on both GSM8K and MATH-500, with high confidence scores ($\mathcal{C}_{cont} > 0.85$). In contrast, GPT-5.1's contamination confidence regresses to baseline levels ($\approx 0.500$). We do not have independent evidence to determine the cause of this difference, and we do not intend this comparison as a claim about GPT-5.1's actual training data.








\end{document}